# From 3D Perception to Safety Reasoning: A Graph-Based Framework for Real-Time Underground Mine Monitoring

Pasindu Ranasinghe[a], Simit Raval[a*], Dibyayan Patra[a], Bikram Banerjee[b], Ismet Canbulat[a]

[a]School of Minerals and Energy Resources Engineering, University of New South Wales, Sydney, NSW, Australia

[b]School of Science, Engineering and Digital Technologies, University of Southern Queensland, Toowoomba, QLD, Australia.

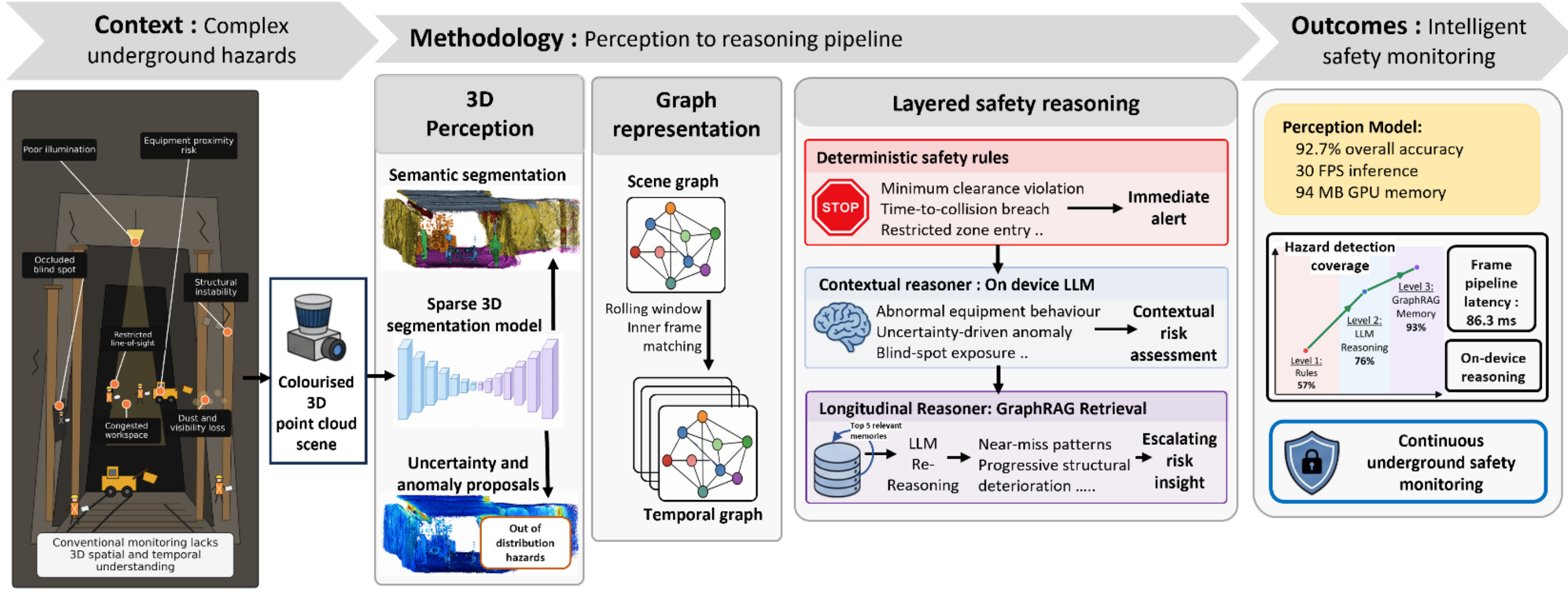

## Abstract

Underground coal mining requires personnel and heavy equipment to operate within shared, confined, and poorly illuminated spaces where hazards such as equipment proximity violations, structural instabilities, and occluded blind spots are difficult to anticipate. Conventional monitoring systems, including fixed cameras and rule-based proximity alerts, can detect predefined events, but lack the three-dimensional scene understanding and contextual memory to identify complex or developing hazards. This paper presents a continuous monitoring framework that converts colourised 3D point clouds into structured and traceable safety reasoning outputs. The framework combines 3D semantic perception, uncertainty-based anomaly detection, rule-based hazard checks, on-device Large Language Model (LLM) reasoning, and graph retrieval-augmented generation (GraphRAG)-based memory analysis to identify immediate hazards and interpret longer-term safety patterns. Scene and temporal graphs serve as the explicit knowledge structure, linking perception outputs across reasoning stages. To overcome the scarcity of labelled underground data, real roadway scans, controlled object placement, and high-fidelity longwall simulation were combined to generate diverse operational and hazard scenarios, while contrastive self-supervised pretraining improved segmentation from limited annotations. The perception model achieved 92.7% overall accuracy at 30 FPS with low memory usage. Across 115 hazard scenarios, rule-based checks achieved 57% coverage, increasing to 76% with contextual LLM reasoning and 93% with memory-based reasoning using historical records. Qualitative results show that uncertainty-derived anomaly signals can support the interpretation of out-of-distribution hazards beyond predefined semantic classes. Overall, graph-based knowledge representation combined with 3D perception and layered safety reasoning provides a practical foundation for intelligent decision support in underground mine monitoring.

---

## 1. Introduction

Underground coal mines remain among the most hazardous industrial workplaces, characterised by confined spaces, harsh environmental conditions, and continuous human–machine interactions that create persistent operational and safety risks. Workers and equipment operate within narrow, cluttered tunnels around roof supports, conveyors, and production machinery, while roadway curvature, structural obstructions, dust loading, low-light conditions, and dynamic machinery movement frequently interrupt line of sight, obscure emerging hazards, degrade sensing reliability, limit depth perception, and reduce overall situational awareness [1-3]. Despite decades of improvement in mining technology, monitoring practices continue to rely heavily on fixed cameras, operator awareness, and periodic inspection — approaches that provide only partial spatial coverage and remain fundamentally limited in their ability to support continuous, three-dimensional assessment of evolving hazards. These limitations become even more critical as underground operations move toward higher production rates, reduced manning, and more automated workflows. Under such conditions, safety monitoring can no longer depend solely on fragmented observations or retrospective reviews; instead, it requires intelligent systems that can continuously monitor the mine environment, interpret complex spatial and temporal interactions, and provide timely, explainable alerts before incidents occur [4]. Prior research in underground mining and automation has consistently highlighted these challenges, particularly in the context of anti-collision monitoring and autonomous operation, where poor visibility, blind spots, and harsh sensing conditions remain fundamental constraints [2]. The severity of this problem is further reflected in recent field incidents. For example, a fatal roof collapse at an underground coal mine in Central Queensland in 2026 resulted in the loss of a worker, while another worker was injured [5]. Such events highlight the continued vulnerability of underground operations to sudden and complex hazards that are difficult to detect with conventional monitoring approaches, emphasising the need for continuous, spatially aware safety systems.

Underground safety is inherently governed by spatial relationships. Many hazardous situations are defined not simply by the presence of objects, but by distance, clearance, trajectory convergence, occlusion, and prolonged occupancy of high-risk zones [6]. For this reason, recent mine automation research has increasingly used LiDAR-based mapping and localisation to capture geometric representations of underground roadways [7]. Studies on underground vehicle localisation and LiDAR-based mapping demonstrate that robust mapping is achievable, while also highlighting persistent challenges in underground perception arising from GPS denial, tunnel aliasing, poor illumination, airborne dust, and feature degeneracy in symmetric roadway geometries. These constraints are not merely implementation challenges; they directly determine the types of monitoring systems that are feasible in underground mines [2]. In contrast, camera-only systems, while widely used in practice, provide limited depth information and are sensitive to lighting, dust, and occlusion, which reduces their reliability in confined underground environments [8]. These limitations motivate the need for representations that preserve both geometric structure and scene context to support autonomous risk assessment beyond visual observation [9].

The colourised 3D point cloud addresses both requirements simultaneously by integrating LiDAR-derived metric geometry with camera-derived appearance cues, preserving both spatial structure and scene context needed for perception and risk assessment [8]. The broader robotics literature has demonstrated the value of camera–LiDAR fusion in dark subterranean environments, ranging from early work on actively illuminated underground void mapping to later approaches for point cloud colourisation using loosely coupled camera systems [10]. These studies demonstrate that multimodal fusion is well-suited to environments with poor lighting and where purely visual methods are often unreliable. In underground mining, colourised 3D sensing is especially attractive because it supports not only reconstruction and localisation, but also the semantic grounding needed for downstream reasoning about personnel, machinery, and infrastructure [8]. Building on our prior research, we have developed an underground LiDAR–camera monitoring platform that produces real-time colourised 3D point clouds in both fixed and mobile configurations [11, 12]. This paper builds upon the existing monitoring platform and extends its scope beyond scene capture. Rather than limiting the system to environmental reconstruction, it establishes a structured perception-to-reasoning pipeline to support autonomous safety monitoring in underground coal mines.

Once a suitable 3D representation is available, the next requirement is semantic understanding of the scene. Over the last decade, point-cloud learning has progressed from direct point-based processing to more efficient sparse convolutional methods that are better suited to large spatial scenes. Foundational architectures such as PointNet [13] and PointNet++ [14] established end-to-end learning on unordered point sets, while later approaches, such as Point Transformer [15], introduced self-attention mechanisms to improve contextual feature learning. For larger and sparser environments, submanifold sparse convolutions and Minkowski convolutional networks [16] made dense 3D inference significantly more practical by avoiding the inefficiencies of dense voxel processing. These developments are directly relevant to underground monitoring, where large-scale point clouds must be processed quickly under restricted compute budgets. At the same time, mining-specific and mining-adjacent studies remain comparatively fragmented. Much of the underground literature has focused on camera-based pedestrian detection and anti-collision monitoring, while LiDAR-based work has mainly addressed mapping, localisation, tunnel inspection, or specific object-identification tasks such as rock-bolt detection, rather than integrated semantic perception frameworks for continuous, real-time safety monitoring [2, 17-20].

Despite these advances, most studies stop at the perception layer, producing detections, maps, or semantic labels without elevating them into explicit safety interpretations [17, 20]. In addition, many contributions are optimised for detection accuracy under controlled conditions without addressing the latency, memory, and real-time throughput requirements of operational deployment. This limitation is evident in practice: a recent roof bolt detection study reports precision up to 94.1%, yet requires approximately 27.5 minutes per scan, making it impractical for continuous real-time monitoring [20]. Furthermore, these approaches are constrained by the realities of mine deployment: annotated underground datasets are scarce, live production panels are difficult to access, hazard-rich data are hard to capture safely, and research devices often face intrinsic-safety and deployment restrictions that make large-scale data collection and validation unusually difficult [21].

For underground deployment, semantic classification alone is not sufficient. Mine environments vary across sites, operating states, and sensing conditions, and predictions that appear plausible at the class-label level may still be unreliable in the presence of occlusion, dust, structural irregularity, or previously unseen configurations. This makes uncertainty estimation and anomaly awareness essential components of a practical monitoring system [22]. Prior work has shown that predictive uncertainty can help identify low-confidence regions and improve reliability in safety-critical perception tasks [23, 24]. In the context of point clouds, these ideas have been extended to uncertainty-aware semantic segmentation and out-of-domain detection in large-scale scans [24]. In addition, unsupervised 3D anomaly localisation approaches have shown that unusual geometric structures can be identified even without exhaustive negative annotation [25]. These capabilities are particularly important in underground mining, where hazardous conditions cannot be fully enumerated during training and may instead emerge as uncertain or anomalous regions rather than predefined semantic classes.

A further limitation of existing underground monitoring research is that, even when perception is successful, the outputs are rarely structured in a form that supports higher-level reasoning. Scene graphs offer a useful alternative by representing entities as nodes and their relationships as edges, transforming dense perception outputs into a relational form that aligns more closely with how safety conditions are defined in practice [26]. In computer vision, 3D semantic scene-graph learning has shown that object relations can be inferred from reconstructed scenes, while in robotics, dynamic scene graphs have been proposed as actionable spatial representations that capture metric and semantic structure across multiple abstraction levels. Kimera extended this idea by showing how dynamic scene graphs can serve as a bridge from SLAM toward more human-meaningful spatial perception [27]. In underground mining, where safety conditions are inherently relational, scene graphs provide a natural abstraction for translating perception outputs into spatially grounded reasoning.

While scene graphs capture the instantaneous spatial configuration of the environment, underground hazards often emerge over time rather than appearing in a single frame. Such hazards may arise through motion, repeated unsafe interactions, proximity escalation, blind-spot occupancy, or persistent anomaly patterns. Standard tracking methods based on inter-frame association, including the Hungarian method [28] and simple 3D tracking baselines, such as AB3DMOT [29], provide practical tools for maintaining temporal consistency. When combined with scene graphs, these tracking methods allow object trajectories, velocities, repeated interactions, and anomaly persistence to be represented in a temporal graph. This temporal dimension is critical in underground monitoring, where a worker repeatedly approaching active machinery, a structural region showing increasing uncertainty, or a near-miss pattern recurring over several frames may only become meaningful through longitudinal interpretation. Although knowledge graphs have been explored for organising mine safety information, these approaches remain largely text-centric and are not connected to real-time 3D perception [30]. As a result, they are limited to static information management rather than dynamic, perception-driven decision-making.

Once perception outputs are represented as structured scene and temporal graphs, large language models become relevant as a higher-level reasoning layer. Rather than using an LLM to directly interpret raw point clouds, the graph provides a compact and auditable input containing object identities, metric relationships, motion descriptors, uncertainty values, and rule-based safety flags. Recent research in robotics and retrieval-augmented reasoning suggests that LLMs can interpret structured relational inputs, combine current observations with retrieved context, and produce more coherent high-level explanations than rule sets alone. Retrieval-augmented generation (RAG) provides a framework for grounding outputs in external evidence, while graph-guided approaches such as KG²RAG [31] and ReGraphRAG [32] demonstrate that incorporating graph structure improves the coherence and relevance of multi-hop reasoning [33]. Emerging coal-mine studies have explored their use for safety assessment, accident-risk analysis, and domain knowledge processing; however, these approaches remain largely text-centric and disconnected from real-time sensing, limiting their suitability for perception-driven monitoring systems in underground environments. This limitation becomes more pronounced under practical deployment constraints. Mine operations impose strict data governance and security policies, requiring sensitive operational data to be processed and stored locally rather than transmitted to external cloud platforms. At the same time, underground installations operate in network-restricted settings where reliable connectivity cannot be assumed [34]. As a result, LLM inference must run entirely on local hardware, limiting the use of large-scale cloud-hosted models and motivating compact, instruction-tuned models that can operate under strict memory and latency constraints [35]. However, the deployment of such LLMs under these conditions remains largely unexplored in the context of real-time underground safety monitoring.

A related research direction has begun to connect 3D data directly with language models. 3D-LLM introduced models that use 3D point clouds and their features for tasks such as 3D captioning, question answering, grounding, navigation, and dialogue, supported by more than 300k generated 3D-language data samples. PointLLM focused more specifically on coloured object point clouds, using 660k simple and 70k complex point-text instruction pairs to align point-cloud features with language-model reasoning and support generative 3D object classification and captioning [36]. ShapeLLM further extended 3D language modelling for embodied interaction by using a ReCon++ point-cloud encoder with multi-view image distillation and 3D visual instruction tuning [37]. These studies demonstrate rapid progress in 3D language understanding, but their training and evaluation remain largely centred on object-level understanding, general 3D question answering, captioning, grounding, and embodied interaction, rather than continuous, safety-critical monitoring of large industrial scenes. This distinction is important for underground mining, where the reasoning layer must operate on large, cluttered, and partially observed scenes while maintaining explicit object identities, metric relations, uncertainty evidence, and auditable outputs.

Taken together, the literature reveals a clear systems gap. Underground mine monitoring research has made meaningful progress in sensing, localisation, reconstruction, and semantic interpretation, yet most contributions remain fragmented and vertically incomplete. Camera-based systems are often limited to 2D observation; LiDAR-based systems tend to focus on mapping and navigation; and point-cloud learning studies typically stop at segmentation or detection. At the same time, scene-graph, temporal reasoning, and LLM-based methods have matured in robotics and related fields and show strong potential for safety monitoring, but their adoption in mining remains minimal, with no integration into real-time mine monitoring. This limitation is consistent with broader mining innovation trends, in which the adoption of advanced digital methods is slower due to high operational risk, regulatory constraints, challenging operating environments, and integration costs in active production settings [38]. As a result, based on the reviewed literature, no fully autonomous framework was identified for continuous hazard detection from real-time sensory inputs.

To address these gaps, this paper makes the following principal contributions:

1. A structured engineering informatics framework for continuous underground safety monitoring:
   The study presents an end-to-end framework that connects colourised 3D sensing, scene understanding, hazard detection, and safety reasoning within one continuous monitoring pipeline.

2. A practical data strategy for restricted underground environments:
   The study combines real roadway scans, controlled object placement, and simulated longwall scenes to create realistic training and evaluation data without requiring repeated access to active production panels.

3. A data-efficient 3D perception approach for limited labelled data:
   The framework applies contrastive self-supervised pretraining before supervised fine-tuning, allowing the segmentation model to learn useful underground scene features from both labelled and unlabelled point clouds.

4. An uncertainty-aware anomaly detection method for identifying hazards beyond predefined semantic classes:
   Per-voxel predictive entropy is used to locate uncertain regions and group them into anomaly proposals, helping identify unseen or out-of-distribution hazardous structures.

5. A structured scene and temporal graph representation for bridging perception outputs and safety reasoning:
   Detected objects and anomaly proposals are encoded as graph nodes and linked across frames to capture distance, proximity, uncertainty, motion, and evolving interactions.

6. A hybrid reasoning layer for current and historical hazard interpretation:
   Rule-based checks detect explicit safety violations, while on-device LLM reasoning and GraphRAG retrieval use current scene context and historical hazard records to interpret complex or recurring risks.

The remainder of this paper is organised as follows. Section 2 presents the proposed methodology, including underground dataset generation, the sparse 3D perception framework, uncertainty-aware anomaly detection, scene and temporal graph construction, the hybrid reasoning layer, and the GraphRAG-based longitudinal reasoning mechanism. Section 3 presents the experimental evaluation and results. Section 4 discusses the implications, limitations, and broader significance of the proposed framework for autonomous underground safety monitoring. Finally, Section 5 concludes the paper and summarises the main contributions and directions for future work.

## 2. Methodology

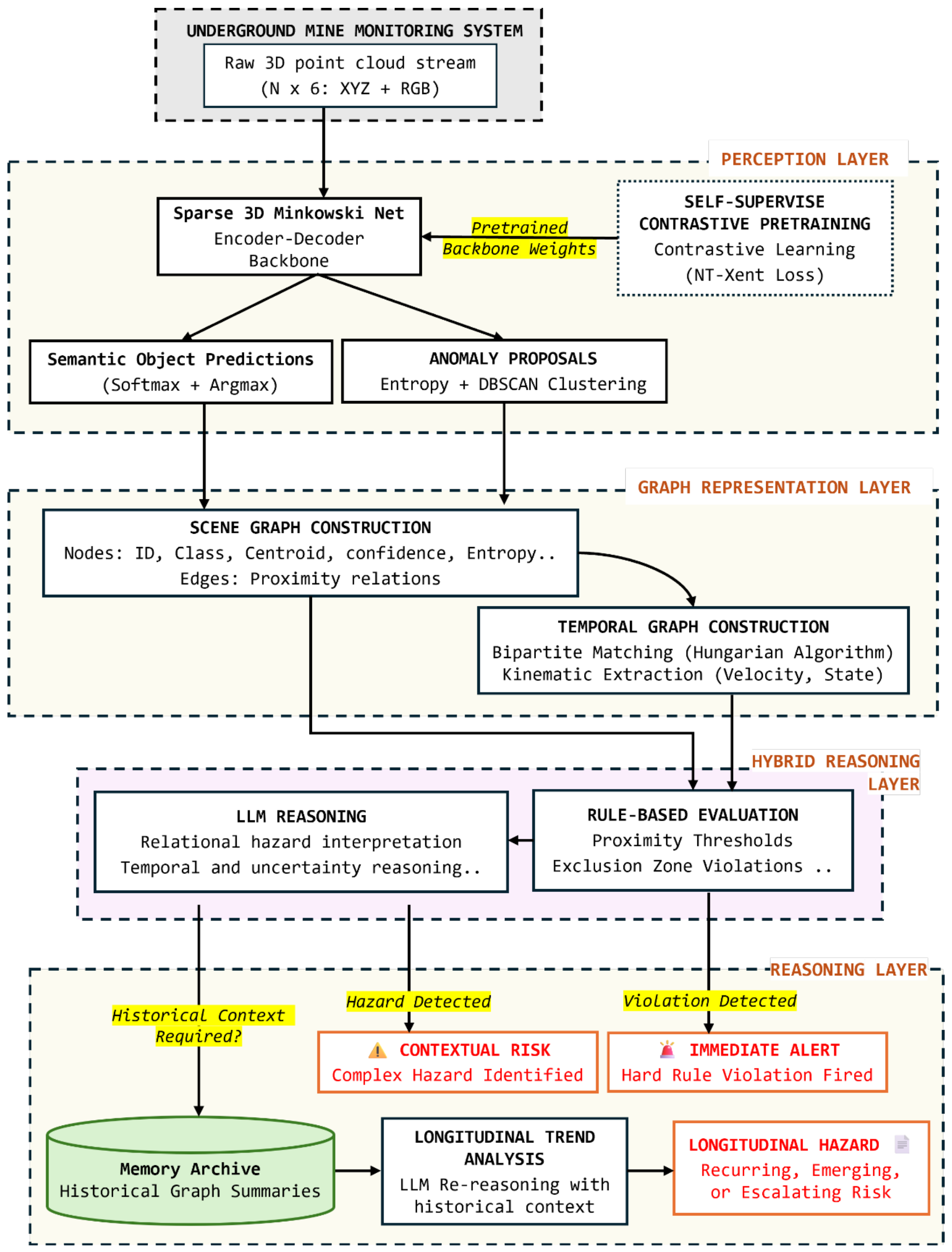

*Figure 1: Overall system architecture of the proposed hybrid framework for underground mine monitoring, detailing the pipeline from sparse 3D perception to graph-based safety reasoning.*

As illustrated in Figure 1, the proposed framework presents a modular safety-monitoring architecture that integrates sparse 3D perception, dynamic scene-graph representation, and Large Language Model (LLM)-based reasoning to support autonomous safety monitoring in underground mining environments. The system operates on real-time multimodal sensor data obtained from a synchronised LiDAR and camera unit. These sensors are integrated within a custom-built device housed in a flameproof enclosure developed in prior work, engineered to satisfy strict safety and certification requirements for hazardous underground mining environments. The system has also been evaluated in multi-device configurations to achieve extended spatial coverage [11, 39]. In this work, the focus is on the monitoring framework built on top of this system, enabling fully autonomous and continuous safety assessment, rather than on the hardware itself.

The captured geometric and visual data are fused to generate colourised 3D point clouds representing the surrounding environment. The point clouds are processed by a sparse 3D neural perception module based on a Minkowski-style convolutional backbone. This module performs dense semantic segmentation to identify key operational entities within the scene, including personnel, shearer components, hydraulic supports, and conveyor structures. To address the scarcity of annotated underground datasets and improve generalisation to unseen environments, the network backbone is first trained using self-supervised contrastive learning and subsequently fine-tuned with supervised semantic labels. During inference, the network predicts per-voxel class probabilities for the segmented scene. Predictive uncertainty is estimated using entropy-based measures derived from these class distributions. Regions with elevated uncertainty are then spatially clustered to generate anomaly proposals, enabling the system to identify out-of-distribution objects or previously unseen hazards. These detected regions are subsequently passed to the higher-level reasoning module for contextual interpretation and safety assessment.

To bridge the gap between raw perception outputs and high-level safety reasoning, detected objects and anomaly proposals are encoded into a structured scene graph at each time step, where nodes encapsulate geometric and semantic descriptors including 3D centroid coordinates, bounding box dimensions, detector confidence, and predictive uncertainty scores, and proximity-based directed edges encode instantaneous spatial relationships between entities. Consecutive scene graphs are then linked through inter-frame identity matching using the Hungarian algorithm to construct a temporal scene graph that explicitly models object trajectories, velocities, kinematic state, and anomaly persistence.

The reasoning pipeline operates in three stages. First, a deterministic rule-based module evaluates each scene graph as it is generated, performing per-frame safety checks such as minimum clearance constraints and exclusion zone violations, and issuing immediate alerts where necessary. In parallel, the temporal graph enables additional rule-based evaluation using short-term motion information. Second, a compact on-device large language model then performs contextual reasoning using the temporal graph, identifying complex relational hazards such as trajectory convergence, blind-spot exposure, and uncertainty-driven anomalies that are not captured by deterministic rules. When the available short-term context is insufficient for reliable interpretation, a GraphRAG mechanism is selectively triggered. In this stage, semantically relevant historical graph summaries are retrieved from a persistent memory archive and combined with the current temporal graph, allowing the language model to perform longitudinal reasoning over extended operational history.

The remainder of this section details the construction of the dataset used to train and evaluate the proposed framework, followed by the architectural design of the Minkowski UNet-based perception backbone and its two-stage training procedure comprising self-supervised contrastive pretraining and supervised fine-tuning, the scene graph generation and temporal graph construction pipeline, the hybrid reasoning layer encompassing the deterministic rule-based module and LLM-based contextual reasoning, and finally the GraphRAG memory architecture and selective retrieval mechanism.

### 2.1 Underground Mining Dataset Generation

Training and evaluating the proposed framework require point cloud data that reflects the geometric complexity, spatial confinement, class diversity, and dynamic operational conditions of real underground mining environments. Unlike surface autonomous systems, where large-scale annotated datasets are widely available, underground mining presents a fundamentally different data acquisition challenge, as confined tunnel geometry, severe illumination constraints, and strict regulatory access requirements make large-scale labelled data collection practically infeasible. To address this, the dataset was constructed from two complementary sources: real field data collected from an operational underground coal mine, providing authentic geometric and structural characteristics, and a high-fidelity simulation environment that generates temporally rich, dynamically varied sequences necessary for training and evaluating the full pipeline. Together, these two sources expose the framework to both the structural authenticity of real underground geometry and the operational diversity of dynamic mining scenarios that cannot be safely or practically captured in the field.

#### 2.1.1 Real Underground Field Data Collection

Field data were collected from an underground coal mine located in the Southern Coalfields region of New South Wales, Australia. The survey corridor extended approximately 1 km along a development roadway, capturing a geometrically representative cross-section of confined underground tunnel conditions, including varying roof profiles, support structures, and conveyor infrastructure. Data acquisition was performed in a development roadway rather than an active longwall face, as access to production panels is tightly restricted by operational safety protocols and intrinsic safety compliance requirements. The roadway environment nonetheless exhibits the geometric complexity, spatial confinement, and environmental characteristics relevant to longwall monitoring applications.

Roadway scans were acquired using a handheld mobile mapping setup comprising a ZEB-REVO laser scanner and an attached camera module, operated at a controlled walking speed of less than 1 $ms^{-1}$. The system continuously recorded synchronised LiDAR, camera, and inertial measurements. Data collection was organised as a sequence of independent 50 m scan sections along the roadway axis. For each section, two scanning passes were performed in opposite directions to increase spatial coverage, minimise occlusions, and improve the accuracy of trajectory estimation. The recorded observations were subsequently processed using the GeoSLAM SLAM algorithm, which reconstructs a geometrically consistent point cloud by fusing LiDAR, visual, and inertial measurements to compensate for platform motion during scanning [40]. This process produced 16 independent point clouds, each representing approximately a 50 m segment of the underground roadway structure.

To introduce operational objects while preserving the authentic underground roadway structure, additional object-level scans were acquired separately using the same sensing system and processing pipeline. Mining vehicles, conveyor components, and personnel were scanned individually to obtain clean, isolated point cloud representations of each object class under real-world underground sensing conditions. These object point clouds were subsequently registered and merged into the corresponding roadway section point clouds at varied spatial positions and configurations, ensuring that each of the 16 scan sections contains realistic object placements with accurate geometric detail, natural occlusion patterns, and authentic sensor characteristics. All 16 point clouds were annotated with voxel- and instance-level labels. Figure 2 shows a representative roadway scan with merged object instances and their annotations.

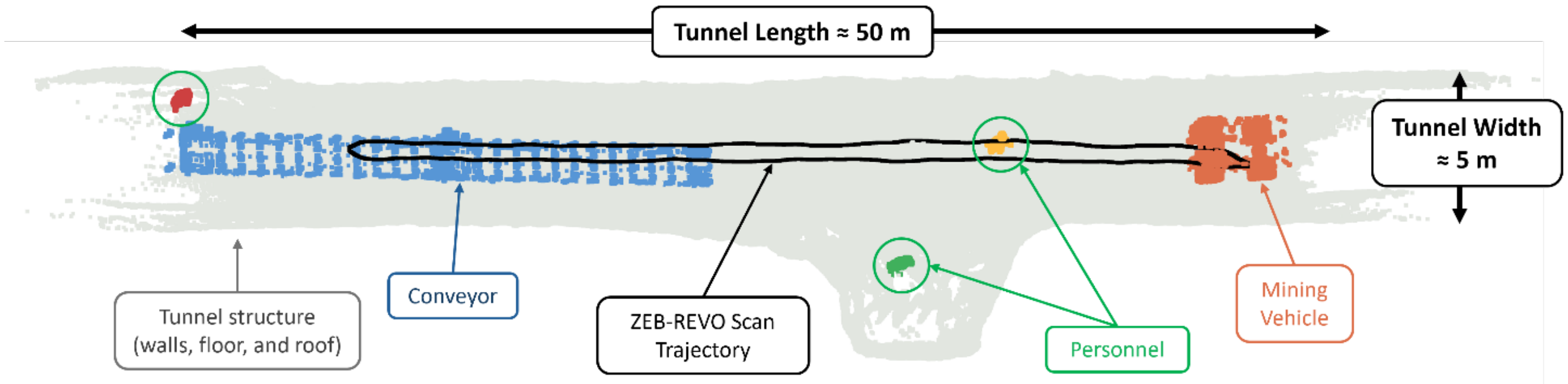

*Figure 2: 50 m roadway scan section showing reconstructed structure (grey), inserted mine objects (coloured), and the scanner trajectory (black).*

### 2.1.2 Simulation-Based Environment

Although field data were collected under representative underground conditions, these recordings alone are insufficient to develop and validate a complete perception-and-reasoning framework for active underground environments. The acquired dataset, while spanning approximately 1 km of roadway, represents a limited temporal window and does not capture the full diversity of operational states, dynamic machinery interactions, or hazardous scenarios that may arise during active coal extraction. Most critically, it lacks sustained temporal sequences involving complex human–equipment interactions, which are critical for safety-oriented monitoring and reasoning.

This scarcity of field data is not incidental but reflects fundamental constraints inherent to underground longwall operations. Longwall faces are active production zones characterised by continuous heavy machinery movement, elevated dust concentrations, structural vibration, and severely limited illumination. These conditions impose strict safety regulations that tightly restrict both physical access and the deployment of research-grade sensing equipment. Electronic devices operating in explosion-risk atmospheres must satisfy intrinsic safety certification requirements, and uncertified LiDAR scanners, mobile mapping platforms, and high-capacity battery systems are generally prohibited without flameproof enclosures and formal regulatory approval. Even when access is feasible, collecting systematically varied data across different operational states, machinery configurations, and hazard conditions would require repeated sensor deployments on live production panels—an approach that is both logistically prohibitive and operationally disruptive. For this reason, simulation was introduced as a controlled extension of the field dataset to provide repeatable safety-critical scenarios for training and evaluation, adding temporal variation, dynamic interactions, and active mining conditions that are difficult to obtain from field data alone. The following simulated longwall and roadway environments were designed to complement the real scan-and-merge dataset while maintaining correspondence with the real roadway data where required.

The simulation framework comprised two environment types constructed in the ROS Gazebo simulator. The longwall panel, illustrated in Figure 3, measured 5 m in width and 18 m in length, with a 3 m roof height consistent with the typical low-mid seam longwall geometries. Structural elements, including hydraulic roof supports, roof canopy, conveyor, and the mining face, were modelled. A shearer model was animated to traverse along the face to emulate cutting operations. Additional dynamic agents representing personnel were introduced with stochastic movement patterns. Controlled lighting conditions were configured between 5–10 lux to replicate dim underground illumination. Stochastically placed geometric clutter and controlled surface perturbations (e.g., small positional offsets and surface noise) were introduced to increase realism and reduce overfitting to idealised geometry.

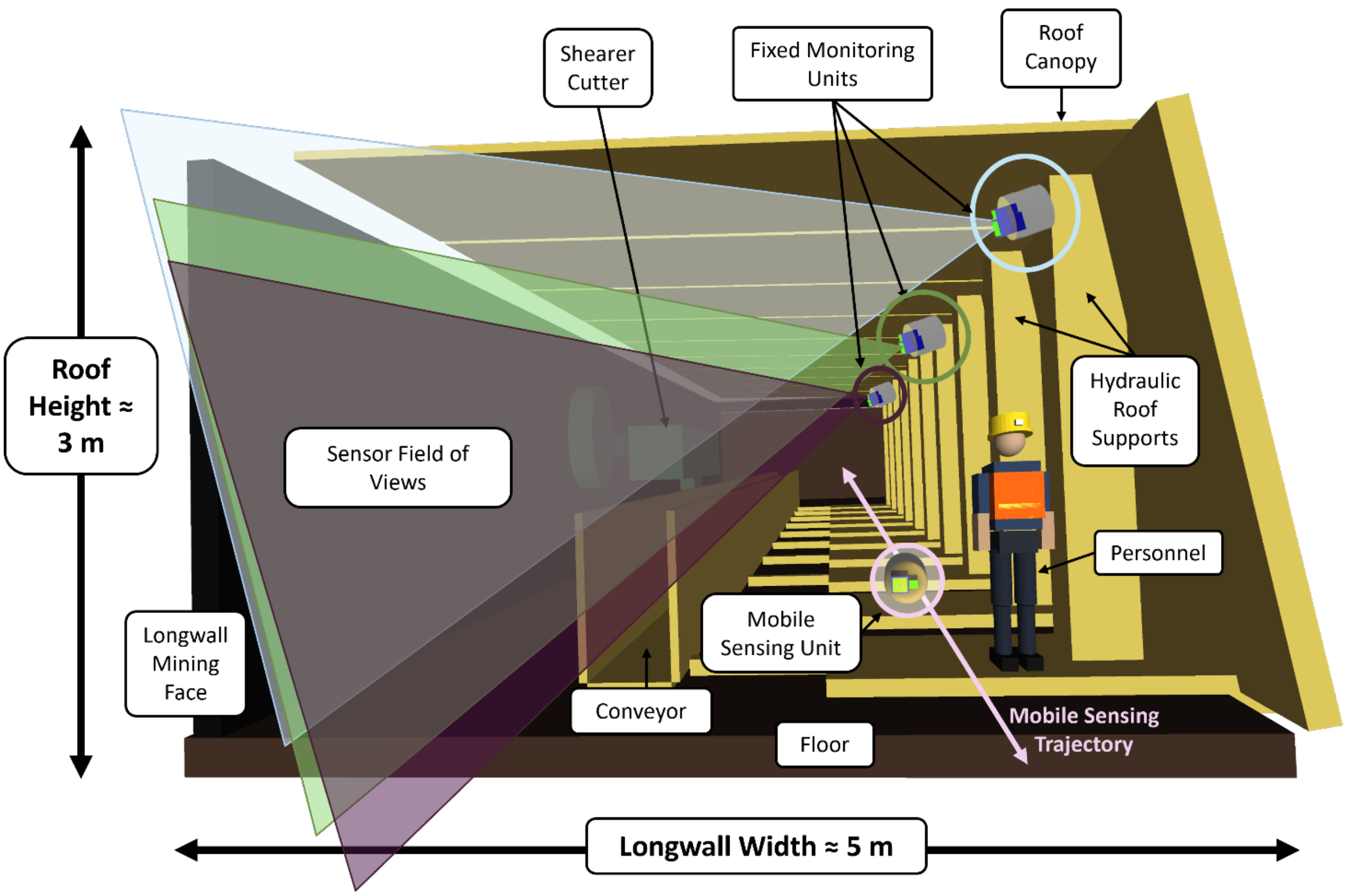

*Figure 3: Simulated longwall environment in ROS Gazebo showing the underground objects. The LiDAR–camera monitoring unit mounted beneath the roof canopy, positioned toward the longwall mining face.*

The tunnel roadway environment was constructed to replicate the confined passage geometry and structural characteristics of the real underground roadway surveyed during field data collection.

The environment was populated with simulated instances of objects representing the same classes used throughout the dataset, including personnel, mining vehicles, and conveyor infrastructure. Object positions and spatial configurations were systematically varied across scans to replicate the object-placement methodology used during real underground data acquisition. This design establishes direct geometric and semantic correspondence between the simulated tunnel environment and the real underground roadway dataset, ensuring consistent exposure to structural features, object diversity, and spatial variability across both synthetic and real tunnel domains. Within these simulated environments, two sensing configurations were implemented for data generation and evaluation, as illustrated in Figure 3: a fixed multi-sensor configuration and a mobile single-sensor configuration.

1. Fixed Multi-Sensor Configuration:
   In the fixed configuration, three identical sensing units were mounted beneath the simulated roof canopy and oriented to maximise coverage of the mining face and operational zone, as shown in Figure 3. The devices were positioned approximately 5 m apart along the tunnel length axis to ensure overlapping yet extended spatial coverage of the longwall face and immediate surroundings. During data collection, controlled variations were introduced to the device mounting orientation, including slight adjustments to the yaw and tilt angles. This was done to simulate realistic installation variability and to generate viewpoint diversity within the dataset. Each sensing unit independently generated a colourised point cloud in its local coordinate frame by projecting LiDAR returns onto synchronised image frames using calibrated intrinsic and extrinsic parameters. The individual colourised clouds were first aligned using calibrated rigid-body transformations and IMU-derived orientation estimates, and subsequently merged into a unified global reconstruction, as illustrated in Figure 4(a). This process produced a dense, fully colourised 3D reconstruction of the monitored longwall region. The overlapping fields of view reduced blind spots, improved coverage of occluded regions, and enhanced overall spatial completeness.

2. Mobile Single-Sensor Configuration:
   In the mobile configuration, a single sensing unit was mounted on a moving platform and programmed to independently traverse both the simulated longwall and tunnel environments. In the longwall environment, the platform progressed along the longitudinal axis of the tunnel as shown in Figure 3, following both predefined inspection routes and stochastic exploratory trajectories. In the simulated tunnel environment, the platform traversed the passage along its longitudinal axis, replicating the handheld mobile scanning methodology used during the real underground field acquisition. During traversal, the sensing unit continuously scanned the confined roadway structure and other elements in the scene (including the shearer, conveyor, and hydraulic supports).

   Throughout platform motion, synchronised LiDAR, camera, and IMU data streams were continuously recorded. The recorded sensor data were processed using the FAST-LIVO2 LiDAR–Inertial–visual SLAM framework to jointly estimate the sensor trajectory and reconstruct the surrounding environment. FAST-LIVO2 performs tightly coupled optimisation by integrating IMU motion estimates with LiDAR scan matching, enabling robust pose estimation even under dynamic motion and partial occlusion [41]. Through incremental state estimation and global map optimisation, the algorithm generates a geometrically consistent 3D reconstruction of the simulated environment together with the corresponding device trajectory. The resulting reconstructions of the active longwall panel and mine roadway environments are illustrated in Figure 4(b).

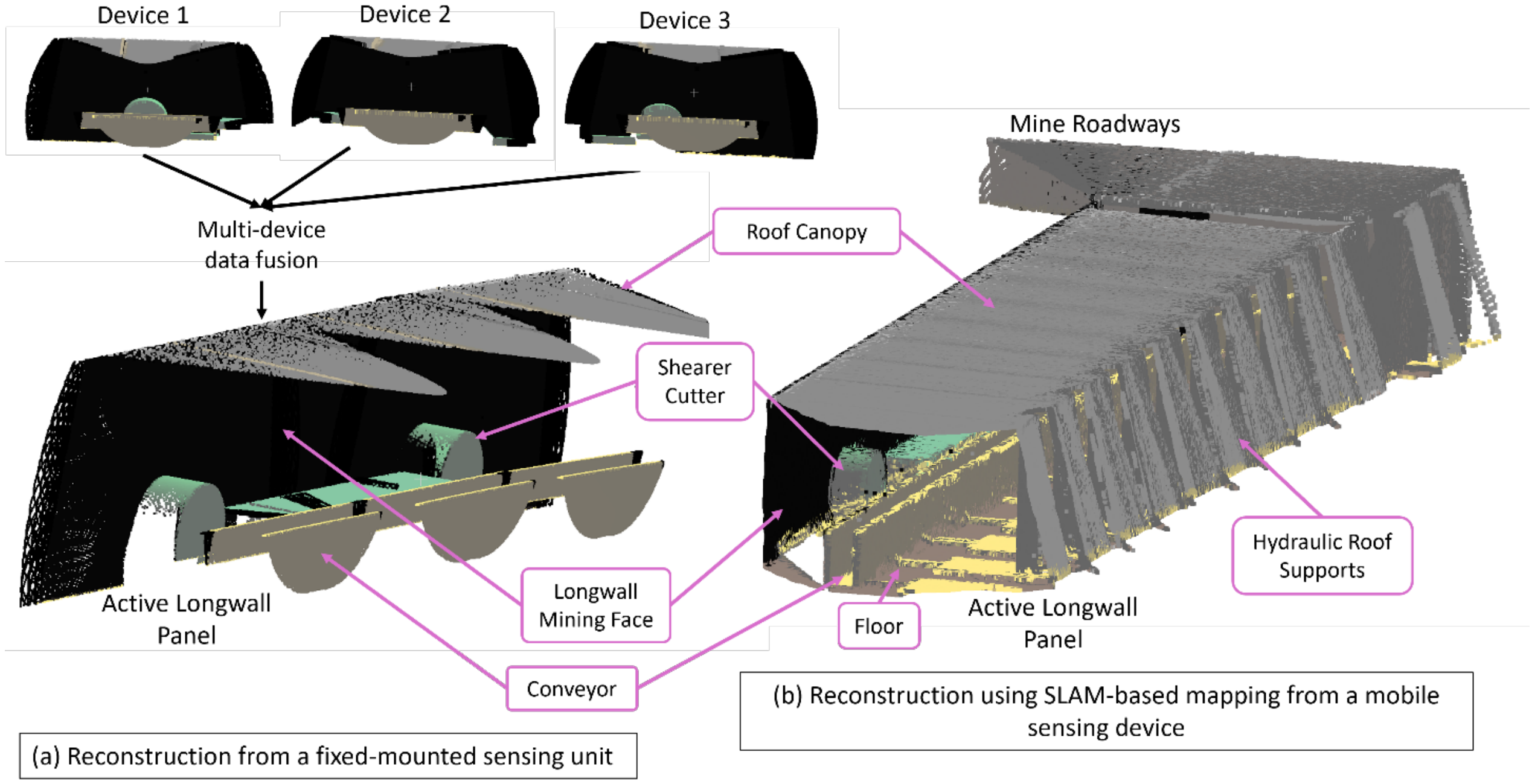

*Figure 4: Multi-sensor data acquisition and reconstruction in underground longwall environments. (a) Fixed multi-device configuration with individual colourised point clouds from Device 1–3 and the resulting merged reconstruction of the active longwall panel. (b) Reconstruction of the underground environment using SLAM-based mapping from a mobile sensing unit, capturing both the active longwall panel and mine roadways.*

### 2.1.3 Dataset Composition

The final dataset comprises data from three acquisition configurations: mobile single-sensor data collected from real underground roadway surveys; fixed multi-sensor data generated within the simulated longwall environment; and mobile single-sensor data acquired across both the simulated longwall and tunnel environments. All point clouds were standardised through voxel downsampling with a uniform voxel size of 0.01 m to ensure consistent spatial resolution across configurations. Table 1 summarises the composition of the dataset.

*Table 1. Dataset summary across acquisition configurations*

| Configuration | Environment | Spatial Coverage | Avg. Points per Cloud (Millions) | Total Point clouds | Annotated Point clouds | Total Annotated Objects |
|---|---|---|---|---|---|---|
| Fixed (3 devices) | Simulated longwall | Longwall face and immediate surroundings | 2.16 | 515 | 25 | 195 |
| Mobile (1 device) | Simulated longwall | Full longwall section | 6.37 | 80 | 15 | 82 |
| | Simulated roadway | Development roadway | 2.87 | 25 | 15 | 76 |
| Mobile (1 device) | Real underground roadway | ~ 1 km development roadway | 3.24 | 16 | 16 | 78 |
| Total | - | - | 2.74 | 636 | 71 | 431 |

Note: The total average points per cloud is calculated as a weighted average using the number of point clouds in each configuration. All point counts are reported in millions.

The three acquisition configurations contribute complementary spatial and structural characteristics to the combined dataset. The fixed multi-sensor configuration provides dense, overlapping colourised observations of the longwall face and immediate operational zone. However, due to the static placement of the sensing units beneath the roof canopy, this configuration is largely limited to the frontal region within the overlapping sensor fields of view. In contrast, the mobile configuration extends spatial coverage by traversing the simulated environments and capturing viewpoints not visible from the fixed sensor positions, including regions surrounding the shearer, structural elements, and confined roadway geometry with varied object placements aligned with the real underground dataset. Together, these configurations provide complementary observations, with the fixed setup contributing dense local coverage and the mobile setup providing broader spatial coverage across the simulated environments.

The underground roadway dataset spans approximately 1 km and consists of 16 spatially distinct 50 m scan sections. These scans preserve the original tunnel structure, support elements, and sensor-noise characteristics. Separately captured point clouds of personnel, mining vehicles, and conveyor components were then inserted into the scan sections at varied locations and orientations to create controlled operational scenes (Figure 2). The resulting merged point clouds combine authentic roadway conditions with realistic object geometry and placement variation. All 16 merged point clouds were annotated with voxel-level and instance-level labels for supervised training and evaluation.

For the simulated datasets, only a subset of point clouds was selected for detailed semantic annotation. In total, 71 point clouds from the combined dataset were annotated using "CloudCompare" software, with voxel and instance-level labels assigned to 9 semantic classes: shearer cutter, mining face, hydraulic supports, roof canopy, conveyor, tunnel walls, mining vehicles, personnel, and floor. The annotated dataset was divided using stratified sampling across all acquisition configurations into 50 training point clouds, 11 validation point clouds, and 10 test point clouds, corresponding approximately to a 70:15:15 split. The remaining 565 unannotated simulated point clouds, together with the 70% training split of the annotated dataset used without labels, were used for self-supervised contrastive pretraining. The validation and test splits were excluded from pretraining to ensure that the evaluation data remained fully unseen. This allowed the backbone to learn from both synthetic and real underground geometry before supervised fine-tuning, while maintaining a clean separation between training and evaluation data.

### 2.2 Semantic Segmentation Model

The perceptual module is implemented as a sparse 3D convolutional neural network based on a customised Minkowski UNet architecture [16], as illustrated in Figure 5. The architecture is adapted for colourised underground point clouds through a six-feature voxel input, a lightweight encoder-decoder backbone, and dual output heads for contrastive pretraining and supervised segmentation. The encoder progressively reduces spatial resolution through strided sparse convolutions to capture high-level geometric context, while the decoder restores spatial resolution using sparse transposed convolutions and symmetric skip connections, preserving fine structural details required for voxel-level prediction. The shared backbone produces a 96-dimensional feature representation for each voxel, providing a compact and discriminative latent space, that serves as the shared interface for both contrastive learning and voxel-level semantic segmentation without architectural modification between stages.

A dual-head design supports a two-stage training strategy to address the scarcity of annotated underground data. During the self-supervised pretraining phase, a contrastive projection head—comprising global sparse average pooling followed by a lightweight multilayer perceptron—is attached to the backbone to learn invariant structural representations from unlabelled data. After pretraining, this head is removed and replaced with a segmentation head consisting of a 1×1×1 sparse convolution that maps backbone features to per-voxel class logits. These logits are used for semantic classification of predefined object categories and for computing voxel-wise predictive entropy, which quantifies model uncertainty and facilitates the detection of anomalous or previously unseen structures.

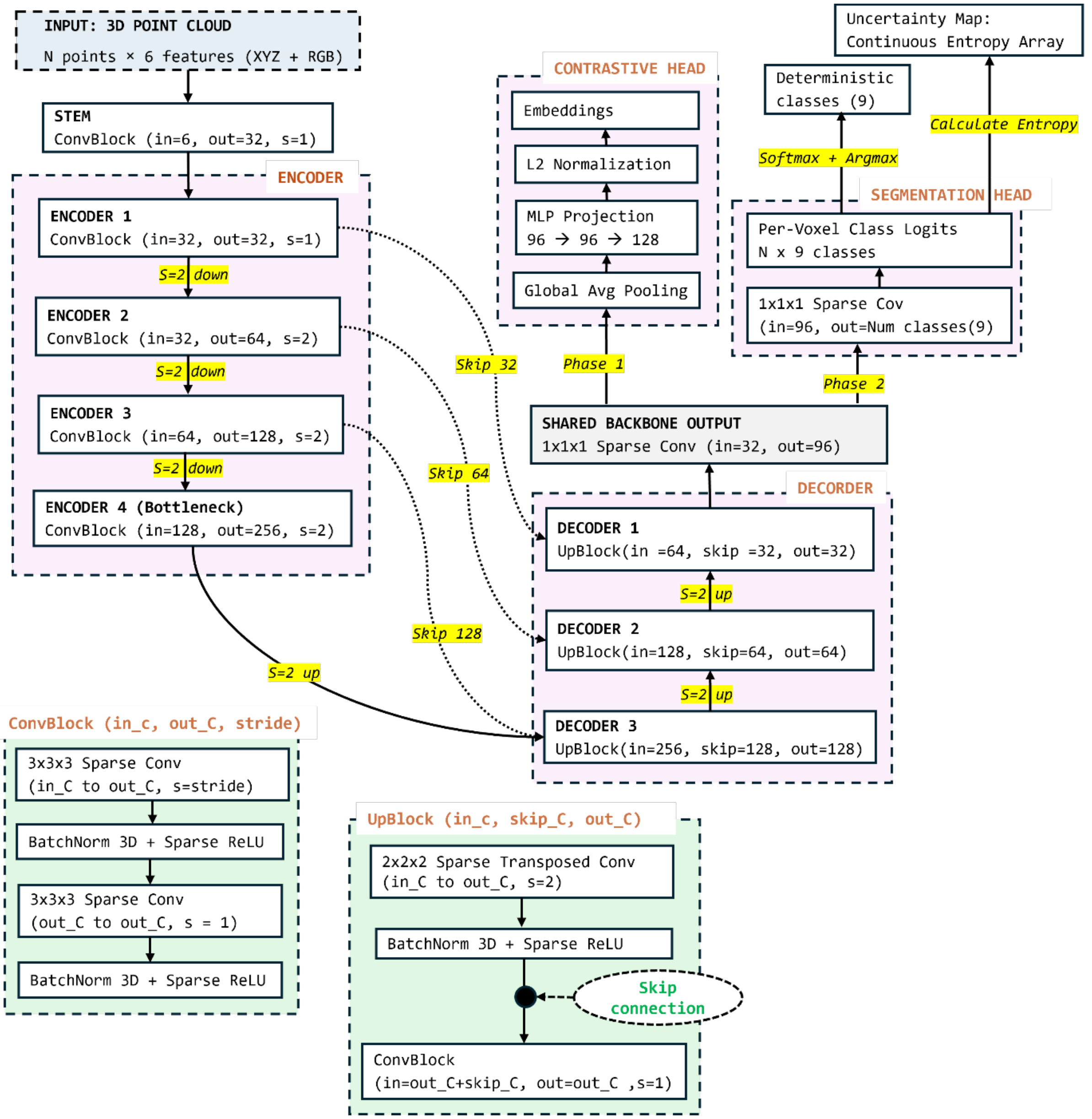

*Figure 5: Architecture of the proposed sparse Minkowski UNet showing the hierarchical encoder–decoder backbone, symmetric skip connections, and dual-head configuration for self-supervised pretraining and supervised segmentation.*

The following subsections describe each architectural component in sequence, beginning with the input stem and progressing through the encoder stages, the bottleneck, the decoder stages, the shared backbone projection, and finally the two output heads and their respective roles within the two-stage training procedure.

### 2.2.1 Input Stem

The input stem converts the voxelised point cloud into an initial feature representation for encoder processing. Each occupied voxel is represented by a six-dimensional input feature vector comprising normalised RGB colour values in the range [0, 1] and spatial coordinates (x, y, z). As shown in Figure 5, this six-feature input is passed to a single ConvBlock, which consists of two sequential 3 × 3 × 3 sparse convolutions, each followed by batch normalisation and Sparse ReLU activation. The ConvBlock projects the input from 6 channels to a 32-channel feature map at the native voxel resolution. No spatial downsampling is applied in the stem, so the original voxel resolution is preserved before hierarchical feature extraction begins in the encoder. The resulting 32-channel feature map is then passed to the encoder, where progressively more abstract spatial and semantic representations are learned.

#### 2.2.2 Encoder

The encoder performs hierarchical feature extraction through four progressive stages that jointly compress spatial resolution while enriching semantic representations. As illustrated in Figure 5, each encoder stage applies a ConvBlock consisting of two sequential 3×3×3 sparse convolution layers, each followed by batch normalisation and ReLU activation. This design deepens feature transformation at a given resolution level without additional spatial reduction, allowing the network to learn richer local geometric patterns before downsampling is applied. The first encoder stage receives the 32-channel feature map produced by the input stem and operates at full spatial resolution using a stride-1 ConvBlock, maintaining voxel density while refining local structural features. From the second stage onwards, stride-2 sparse convolutions progressively reduce spatial resolution by a factor of two while increasing channel dimensionality, enabling the network to capture broader contextual information. As a result, the feature dimensionality expands from 32 channels in the first stage to 64 and 128 channels in subsequent stages, ultimately reaching a 256-channel bottleneck. This bottleneck encodes the highest-level geometric abstraction of the scene and provides the global context required for accurate reconstruction in the decoder.

#### 2.2.3 Decoder

The decoder restores voxel-level spatial resolution through three upsampling stages corresponding to the three stride-2 encoder stages. Each stage begins with a 2×2×2 sparse transposed convolution with a stride of 2, which doubles the spatial resolution of the voxel grid. The upsampled feature map is then concatenated with the corresponding skip connection from the symmetric encoder stage, reintroducing fine-grained spatial information that was compressed during encoding. The fused features are subsequently refined using a ConvBlock identical to those employed in the encoder, consisting of two sequential 3×3×3 sparse convolutions followed by batch normalisation and ReLU activation. As spatial resolution increases across the decoder stages, channel dimensionality is progressively reduced from 256 to 128 and 64 channels, ultimately producing a 32-channel full-resolution feature map, which is then forwarded to the shared backbone projection layer before branching into the task-specific output heads.

#### 2.2.4 Shared Backbone Output

The 32-channel full-resolution feature map produced by the final decoder stage is projected to a fixed 96-channel representation using a 1×1×1 sparse convolution. This projection expands the decoder output into a higher-dimensional latent space that serves as the shared interface for the subsequent task-specific heads. The resulting 96-dimensional per-voxel feature representation provides sufficient expressive capacity to support both contrastive learning during the self-supervised pretraining stage and voxel-level semantic classification during supervised fine-tuning, enabling both tasks to share a common backbone without architectural modification.

### 2.2.5 Segmentation Head

The segmentation head maps the 96-dimensional per-voxel feature representation to semantic class logits using a 1×1×1 sparse convolution, producing a 9-dimensional logit vector for each active voxel corresponding to the predefined semantic classes. Since the convolution operates independently at each voxel location, spatial resolution and voxel connectivity remain unchanged, preserving strict spatial alignment between input coordinates and predicted labels. The resulting logits are converted to class probabilities using the softmax function. The maximum-probability assignment yields deterministic semantic labels for each voxel, enabling per-voxel classification across the nine predefined semantic classes. Beyond deterministic classification, voxel-wise predictive entropy is estimated from the full probability distribution, without introducing any additional inference cost. Its computation and use for anomaly region localisation are described in detail in Section 2.2.9.

### 2.2.6 Contrastive Projection Head

During the self-supervised pretraining stage, the segmentation head is replaced with a contrastive projection head attached to the shared backbone output. Global sparse average pooling aggregates the 96-dimensional per-voxel feature map into a compact scene-level descriptor, reducing the spatial dimension while preserving the global feature distribution. This descriptor is then processed by a lightweight two-layer multilayer perceptron (MLP) that projects the representation from 96 to 128 dimensions. The resulting embedding is subsequently L2-normalised before computing the contrastive loss. The projection head is used exclusively during the pretraining stage to learn geometrically invariant structural representations from unlabelled point clouds. After pretraining, the projection head is removed, and the learned backbone weights are retained to initialise the segmentation network for supervised fine-tuning.

### 2.2.7 Self-Supervised Pretraining via Contrastive Learning

The shared backbone is optimised using contrastive learning to acquire geometrically invariant structural representations. Pretraining is performed on a combined pool comprising 615 point clouds, including 565 unlabelled simulated point clouds and the 50 point clouds in the training split of the annotated dataset, used in a label-free manner. For each input point cloud, two independent augmented views are generated. The augmentation pipeline includes random rotation about the vertical axis within ±30°, uniform scaling in the range 0.9–1.1, Gaussian jitter noise with a standard deviation of 0.01 m, and random point dropout of up to 10%. These transformations alter the geometric appearance of the scene while preserving its semantic structure. As a result, the network is encouraged to learn structural representations that are invariant to common sensor noise, viewpoint changes, and partial occlusions.

Contrastive optimisation is performed using the normalised temperature-scaled cross-entropy loss (NT-Xent) [42]. For a mini-batch containing $B$ point clouds, two augmented views are generated for each sample, resulting in a total of $2B$ embeddings. Given a positive pair of embeddings $(\mathbf{z}_i, \mathbf{z}_j)$ derived from the same original point cloud, the loss is defined as:

$$L_{i,j} = -\log \frac{\exp(\mathrm{sim}(z_i, z_j)/\tau)}{\sum_{k=1}^{2B} 1_{[k \neq i]} \exp(\mathrm{sim}(z_i, z_k)/\tau)}$$

where $\mathrm{sim}(z_a, z_b) = z_a^\top z_b$ denotes cosine similarity between two L2-normalised embeddings, $\tau$ is the temperature hyperparameter controlling the sharpness of the similarity distribution, and the indicator term $\mathbf{1}_{[k \neq i]}$ excludes self-comparisons from the denominator. The total loss is computed by averaging $L_{i,j}$ across all positive pairs in the batch.

The temperature parameter was set to $\tau = 0.1$, providing a balance between discriminative separation and optimisation stability [42]. The backbone was trained using the Adam optimiser [43] with an initial learning rate of $1 \times 10^{-3}$, decayed using a cosine learning-rate schedule [44] over 100 epochs. A batch size of 16 point clouds was selected to balance GPU memory constraints while maintaining sufficient negative samples per batch for effective contrastive learning.

#### 2.2.8 Supervised Fine-Tuning

Following self-supervised pretraining, the contrastive projection head is removed and replaced with the segmentation head described in Section 2.2.5. The pretrained backbone weights are retained to initialise the supervised model, transferring the geometric structural priors learned from unlabelled data into the supervised segmentation stage. The segmentation network is then fine-tuned for voxel-level semantic classification on the annotated dataset. Given an input sparse point cloud with N active voxels, the segmentation head produces per-voxel logits $\mathbf{Z}_i \in \mathbb{R}^C$, where $C$ = 9 denotes the number of predefined semantic classes. These logits are converted to class probabilities using the softmax function:

$$p_{i,c} = \frac{\exp(Z_{i,c})}{\sum_{k=1}^{C} \exp(Z_{i,k})}$$

where $p_{i,c}$represents the predicted probability that voxel $i$ belongs to the class $c$. The network is trained using the standard cross-entropy loss ($L_{seg}$):

$$L_{seg} = -\frac{1}{N} \sum_{i=1}^{N} \sum_{c=1}^{C} y_{i,c} \log(p_{i,c})$$

where $y_{i,c}$ is the ground-truth one-hot label for voxel $i$.

Fine-tuning was performed using the Adam optimiser [43] with a cosine annealing schedule [44] over 100 epochs. A batch size of 8 point clouds was used during supervised training, with early stopping based on validation loss to mitigate overfitting given the limited annotated data. To preserve the structural priors learned during contrastive pretraining, differential learning rates were applied during the initial fine-tuning stage: the pretrained backbone was updated with a learning rate of $1 \times 10^{-4}$, while the segmentation head was trained with a learning rate of $5 \times 10^{-4}$.

This allowed the segmentation head to adapt rapidly to the labelled classes while the backbone weights changed more conservatively.

### 2.2.9 Uncertainty Map Calculation

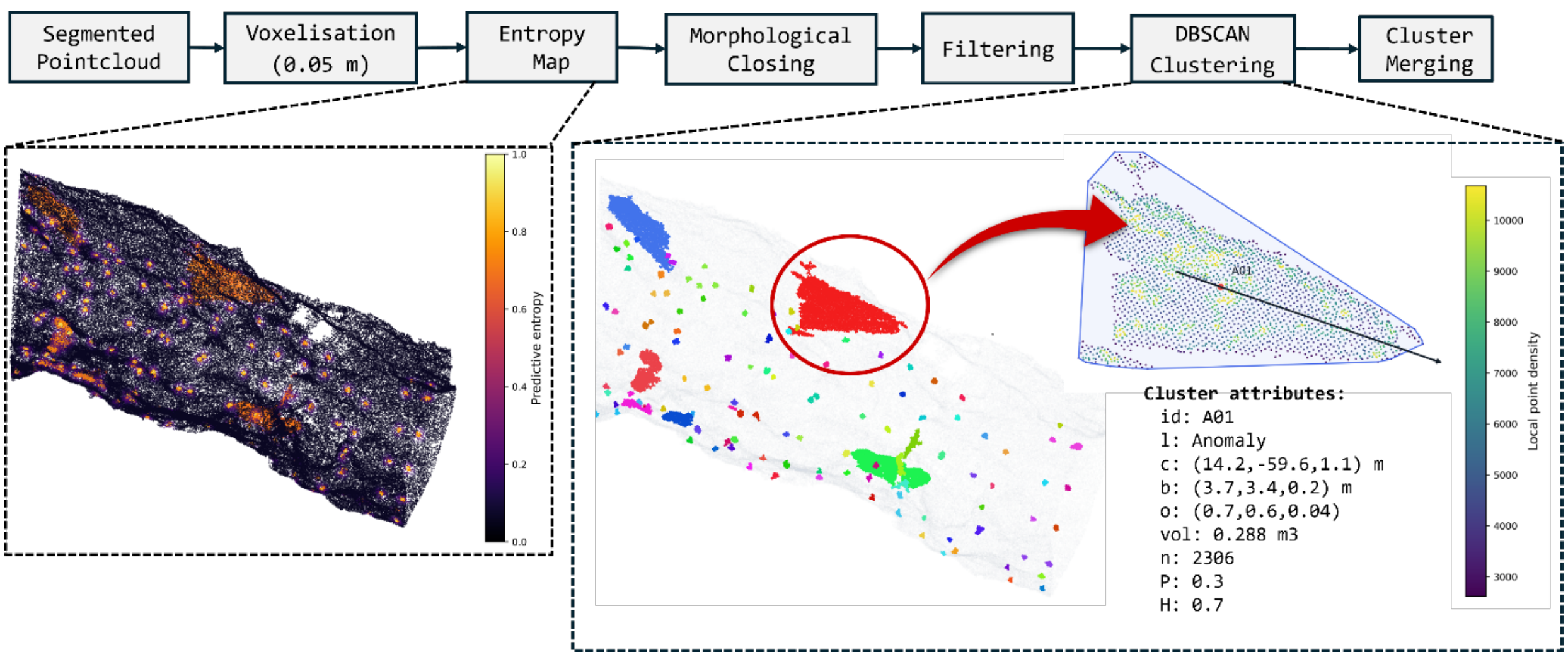

*Figure 6. Uncertainty-driven anomaly extraction from an underground roadway point cloud. Top: processing pipeline. Bottom: entropy map, DBSCAN clusters, and a representative anomaly cluster with associated graph attributes (details in Section 2.3).*

In addition to deterministic semantic classification, spatial uncertainty is used to localise candidate anomaly regions within the scene. Uncertainty is quantified from the per-voxel class probability distribution produced by the segmentation head using predictive entropy. For voxel $i$, with class probabilities $p_{i,c}$ over $c$ semantic classes, the entropy is defined as $(H_i)$:

$$H_i = -\sum_{c=1}^{C} p_{i,c} \log(p_{i,c})$$

Low entropy corresponds to confident predictions, whereas higher entropy indicates ambiguous predictions distributed across multiple classes. In underground environments, such uncertainty commonly arises due to occlusions, dust, sensor noise, sparse observations, and previously unseen structures, making entropy a reliable indicator of potential anomaly regions. Voxels exceeding a predefined entropy threshold of 0.35 (approximately 15% of the maximum entropy) are identified as high-uncertainty locations where no dominant semantic class is predicted. The resulting entropy map highlights these regions, as illustrated in the bottom-left panel of Figure 6.

However, the thresholded uncertainty map often contains fragmentation, small gaps, and isolated noise. To address this, a filtering stage is applied. A 3D morphological closing operation (single iteration) is first used to bridge local discontinuities, followed by the removal of small, connected components below a minimum size threshold (20 voxels). The filtered voxels are then grouped using the Density-Based Spatial Clustering of Applications with Noise (DBSCAN) algorithm [45] with ε=0.10 m, producing initial anomaly clusters, as illustrated in the bottom-right panel of Figure 6. To obtain coherent regions, clusters are further merged when their boundary separation was below 0.15 m, their centroid separation was below 0.20 m, and their principal orientation differed by less than 15°. Each resulting cluster is treated as a candidate anomaly region and added to the scene graph as an anomaly node, alongside the existing semantic object instances, as described in Section 2.3.

### 2.3 Scene Graph Generation

At each time step, the geometric and semantic predictions produced by the perception backbone are abstracted into a structured scene graph, providing a compact relational representation of the monitored environment. The spatial configuration at time $t$ is represented as a graph $G_t = (\mathcal{V}_t, \mathcal{E}_t)$, where $\mathcal{V}_t$ denotes the set of object nodes and $\mathcal{E}_t$ represents the spatial relations between them. This abstraction converts dense voxel-level predictions into a structured representation that preserves object identity, geometry, and contextual interactions, making it suitable for downstream reasoning as illustrated in Figure 7.

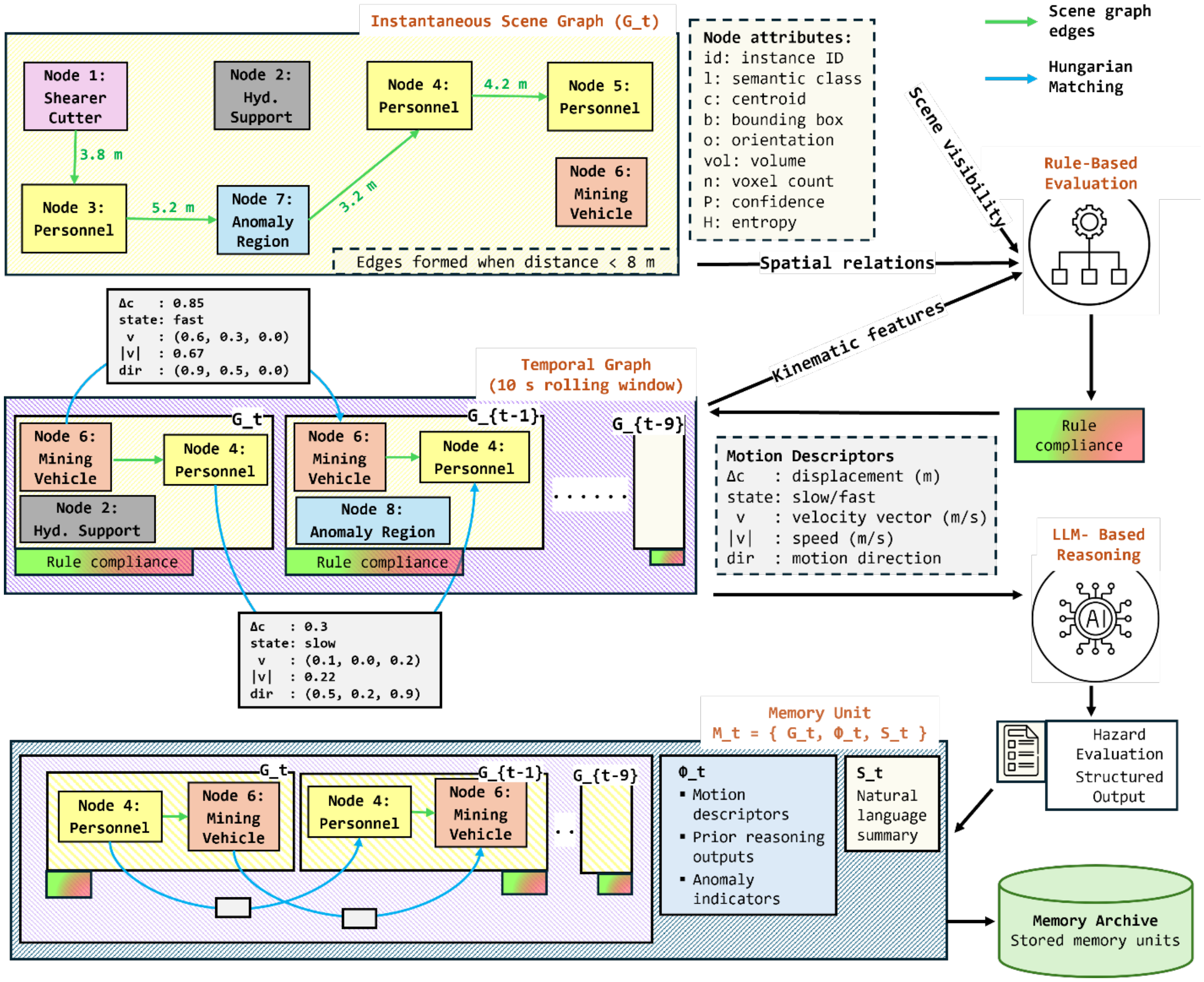

*Figure 7: Scene graph construction, temporal graph tracking, and memory unit generation pipeline. Nodes represent detected objects with attributes $v_i$, and edges encode spatial relations within an 8 m threshold. Consecutive scene graphs are linked via inter-frame association over a 10 s rolling window to extract kinematic descriptors. Each reasoning step produces a memory unit $M_t = \{G_t, \Phi_t, S_t\}$, which is stored for longitudinal retrieval.*

Each distinct object instance extracted from the segmentation output and anomaly proposal clustering is represented as a node $v_i \in V_t$ in the scene graph at time $t$. Each node is parameterised as:

$$v_i = (\mathrm{id}_i,\ \mathrm{l}_i,\ c_i,\ b_i,\ o_i,\ \mathrm{vol}_i,\ n_i,\ \overline{p_i},\ \overline{H_i})$$

where $\mathrm{id}_i$ denotes the unique instance identifier assigned during clustering, $\mathrm{l}_i$ represents the semantic class predicted by the segmentation model, $\mathbf{c}_i \in \mathbb{R}^3$ denotes the 3D centroid, $\mathbf{b}_i \in \mathbb{R}^3$ represents the axis-aligned bounding box dimensions, $\mathrm{vol}_i$ represents the estimated spatial volume, $n_i$ represents the voxel count, $\bar{p}_i$ denotes the mean softmax confidence, and $\bar{H}_i$ is the mean predictive entropy across the instance. The orientation vector $\mathbf{o}_i$ is estimated using principal component analysis (PCA) on the voxel coordinates of the instance. It is defined as the first principal component, corresponding to the dominant eigenvector of the covariance matrix, which captures the primary direction of spatial extent. Expressed in the global coordinate frame, this representation provides directional information that complements the centroid and bounding box geometry.

Edges in $\mathcal{E}_t$ encode instantaneous spatial relationships between object node pairs. For each ordered pair $(v_i, v_j)$, the Euclidean distance between their centroids is computed as $d_{ij} = \| \mathbf{c}_i - \mathbf{c}_j \|_2$. An edge $e_{ij}$ is created when $d_{ij} \leq 8$ m. This threshold reflects the spatial scale of the longwall panel and captures interactions within the immediate operational region. At typical underground movement speeds of around 1.5–2 m/s, an 8 m separation corresponds to approximately 4–6 seconds of potential interaction time, providing a practical margin for early detection of proximity between personnel and machinery before critical distances are reached. The Euclidean distance $d_{ij}$ is stored as a continuous edge attribute that represents the metric proximity between connected objects.

**2.4 Temporal Graph Construction**

Machinery continuously traverses the longwall face, personnel move through the operational zone, and spatial relationships between entities evolve over time in ways that cannot be adequately characterised from a single frame. To capture object persistence, motion, and trajectory interactions across consecutive observations, scene graphs from successive time steps $\{G_t, G_{t+1}, \dots, G_{t+n}\}$ are linked through inter-frame identity association to form a unified temporal graph representation. This representation is maintained as a rolling 10-second spatiotemporal window, updated at each frame by appending the current scene graph and discarding outdated observations. This window is sufficient to capture short-term motion patterns such as approach trajectories and velocity changes, while remaining compatible with the context limits of the on-device language model used for downstream reasoning; its suitability is evaluated further in the Results section.

Inter-frame association is formulated as a bipartite matching problem between node sets $\mathcal{V}_t$ and $\mathcal{V}_{t+1}$. The Hungarian algorithm is used to compute optimal assignments based on a cost matrix that combines centroid displacement and semantic consistency [46]. The association cost between the node $i$ at time $t$ and node $j$ at time $t+1$ is defined as:

$$\mathcal{C}_{ij} = \| \mathbf{c}_i^t - \mathbf{c}_j^{t+1} \|_2 + \lambda \cdot \mathbf{1}[\ell_i \neq \ell_j]$$

where $\| \mathbf{c}_i^t - \mathbf{c}_j^{t+1} \|_2$ denotes the Euclidean centroid displacement between the candidate pair, and $\mathbf{1}[\ell_i \neq \ell_j]$ is an indicator function for semantic label mismatch. A large penalty parameter $\lambda = 1000$ is applied to strongly discourage cross-class identity switches during matching. In addition, assignments are constrained by a maximum inter-frame displacement threshold of 1.0 m; candidate matches exceeding this distance are rejected, and the unmatched node is initialised as a new track. The gating threshold is selected to accommodate typical inter-frame object displacement at the given acquisition rate, while preventing incorrect cross-object associations. Once associations are established, kinematic descriptors are extracted over time. For each tracked object, displacement is computed as $\Delta \mathrm{c} = \mathrm{c}^{t+1} - \mathrm{c}^t$ and velocity is estimated as $v = \Delta \mathrm{c} / \Delta t$ using timestamp metadata. Movement states are categorised as stationary (displacement < 0.2 m), slow (0.2–0.5 m), or fast (> 0.5 m) per frame interval. For moving objects, motion descriptors, including track identity, velocity vector, speed, movement state, and direction of motion are generated and added to the temporal graph, as shown in Figure 7. The resulting spatiotemporal representation is then provided to the downstream hybrid reasoning module for safety assessment.

### 2.5 Hybrid Contextual Reasoning and Hazard Detection

While the temporal graph encodes the geometric and kinematic state of the underground environment, translating this structured representation into actionable safety assessments requires a further layer of interpretation. Underground safety scenarios frequently involve a mixture of deterministic threshold violations and contextually evolving risk conditions that cannot be captured by geometric rules alone. To address this, the framework introduces a hybrid reasoning layer that combines a deterministic rule-based pre-evaluation system with a compact on-device large language model. This dual-layer design enables real-time enforcement of hard safety constraints while providing contextual interpretation for complex relational hazard scenarios beyond predefined thresholds.

#### 2.5.1 Deterministic Rule-Based Pre-Evaluation

The first reasoning layer is a deterministic rule-based module that acts as a near-zero-latency safety gate. It is applied at every update step as new observations are incorporated into the temporal graph. The module evaluates five explicit hazard conditions drawn from three sources: spatial and geometric conditions are evaluated from the current scene graph, motion-dependent conditions are derived from the temporal graph, and visibility degradation is assessed from the colourised point cloud. These rules cover static proximity violation, predictive collision risk, blind-spot occupancy, operational-zone congestion, and severe visibility degradation.

The first rule detects static proximity violations. When an equipment node is classified as active, a static proximity hazard alert is triggered if the Euclidean centroid distance satisfies $d_{ij} \leq 1.5$ m, indicating that personnel have entered the crush hazard zone of operating machinery. The second rule extends this check by evaluating motion-based collision risk. A predictive collision alert is generated when active equipment is moving toward a personnel node and the estimated Time-To-Collision (TTC) satisfies $\mathrm{TTC}_{ij} \leq 3.0$ s. The TTC is defined as:

$$\text{TTC}_{ij} = \frac{d_{ij} - d_{safe}}{\| \mathbf{v}_j - \mathbf{v}_i \| \cdot \cos\ \theta}$$

where $d_{\text{safe}} = 1.5$ m denotes the minimum safe separation distance, $\mathbf{v}_i$ and $\mathbf{v}_j$ denote the velocity vectors of the personnel and equipment nodes, respectively, and $\theta$ represents the angle between the relative velocity vector and the centroid-to-centroid direction vector. TTC is only evaluated when $\cos\ \theta > 0$, indicating approaching motion; diverging pairs are excluded.

The third rule detects blind-spot occupancy based on the relative position of personnel with respect to the equipment heading or operational facing direction. For each active equipment node $v_j$, the facing direction is represented by the orientation vector $\mathrm{o}_j$, obtained from the PCA-estimated principal axis and aligned with the equipment motion direction where available. The angular offset between the personnel node $v_i$and the equipment facing direction is computed as:

$$\phi_{ij} = \cos^{-1}\left(\frac{(\mathbf{c}_i - \mathbf{c}_j) \cdot \mathbf{o}_j}{\| \mathbf{c}_i - \mathbf{c}_j \| \ \ \| \mathbf{o}_j \|}\right)$$

where $\mathrm{c}_i$ and $\mathrm{c}_j$ denote the centroids of the personnel and equipment nodes, respectively. A blind-spot alert is triggered when $\phi_{ij} > 135°$and $d_{ij} \leq 3.0$m, indicating that the personnel node lies within the rear sector of the equipment and remains close enough to represent an operational safety concern. This captures scenarios where personnel are positioned behind or on the rear side of active equipment, outside the assumed forward awareness sector, and therefore may not be readily visible during equipment movement or operation.

The fourth rule detects operational-zone congestion by evaluating personnel occupancy within the immediate vicinity of active equipment. For each active equipment node $v_j$, all personnel nodes within an operational radius of $r = 4.0$ m are counted, and the local personnel density is computed as:

$$\rho_j = \frac{n_j}{\pi r^2}$$

where $n_j$ denotes the number of personnel nodes within a radius $r$ of the equipment node $v_j$. A congestion alert is triggered when $\rho_j$ exceeds a predefined density threshold of 0.06 personnel/m², indicating that the operational zone around active equipment is occupied beyond a safe level [47]. This provides a conservative rule for confined underground roadways, where multiple personnel working close to active machinery can reduce clearance, increase occlusion, and elevate interaction risk

The fifth rule monitors scene visibility by computing the mean RGB intensity across active voxels in the colourised point cloud, averaged over a rolling 3-second temporal window. An alert is triggered when this rolling mean falls below a threshold of 25 out of 255, indicating severe illumination degradation that reduces situational awareness across the monitored zone.

All alert types are immediately propagated to the safety output layer for operator notification and recorded as edge attributes on $e_{ij}$ within the current scene graph. This ensures that the downstream reasoning stage has access to all active deterministic hazard indicators when analysing the scene context. Additional deterministic rules capture other operational violations, such as personnel or equipment entering restricted coordinate zones defined by the mine safety protocol. While this deterministic layer guarantees strict compliance with predefined safety thresholds in real time, it is inherently limited in its ability to capture contextual hazards that require higher-level or longer-horizon interpretation beyond explicit rule definitions. These limitations motivate the integration of a language model-based reasoning layer capable of interpreting higher-level spatial and temporal relationships.

### 2.5.2 LLM-Based Contextual Reasoning

The deterministic rule-based layer and the LLM-based reasoning module operate in parallel within the hybrid framework. The rule-based system remains the primary mechanism for detecting predefined hazards and flags all threshold-based violations with zero latency. The LLM module does not re-evaluate these detections; instead, it focuses on scenarios not captured by predefined rules or that require additional contextual interpretation. In particular, the LLM analyses relational and temporal patterns to identify hazards beyond explicit thresholds, such as abnormal equipment behaviour, gradual trajectory convergence, and uncertainty-driven anomalies. It can also detect emerging risk conditions in frames where no rule-based alert is triggered. This complementary design preserves deterministic detections while enabling identification of previously unseen or context-dependent hazards, thereby improving overall system awareness.

LLM selection was guided by deployment constraints of underground edge systems, including limited GPU memory (≤ 16 GB), restricted power availability, and real-time requirements. These constraints render large-scale foundation models impractical for on-device inference, necessitating compact instruction-tuned models capable of structured reasoning within the available hardware budget. Candidate models were initially screened based on their benchmark performance in structured reasoning, instruction-following, context capacity, and inference efficiency. Three models were shortlisted for detailed evaluation: Qwen2.5-3B-Instruct [48], Phi-4-mini-instruct [49], and DeepSeek-R1-Distill-Qwen-1.5B [50]. Based on the evaluation of simulated hazard scenarios, Qwen2.5-3B-Instruct was selected as the reasoning module. A comprehensive comparison is provided in the Results section.

To enable this interpretation, the temporal scene graph is serialised into a structured textual representation and provided to an on-device large language model. The serialisation converts the graph into a constrained natural-language schema in which each node is represented by its instance identifier, semantic label, 3D centroid coordinates, bounding box dimensions, principal-axis orientation, velocity vector, movement-state classification, mean softmax confidence, and mean predictive entropy. Each edge is represented by the identifiers of connected nodes, their Euclidean distance, geometric relational predicates, and any deterministic safety flags generated during the rule-based stage. This representation scales with object count rather than voxel density, ensuring that prompt length remains bounded and computationally feasible under underground edge deployment constraints.

The LLM operates under a structured prompt template that constrains outputs to predefined schema fields, requiring structured risk assessments that explicitly reference object identifiers and relational evidence within the serialised graph. This avoids free-form narrative generation and ensures that each output remains directly traceable to specific graph elements, preserving interpretability and auditability. Accordingly, the model produces a structured hazard assessment comprising identified risk conditions, the object identifiers involved, severity classification, and a brief explanation based on the graph representation.

Consistent with the hybrid reasoning design, deterministic safety alerts remain authoritative and are executed independently of language model inference, ensuring that critical safety violations are not subject to probabilistic reasoning or additional latency.

### 2.6. Longitudinal Analysis via Graph Retrieval-Augmented Generation (GraphRAG)

The hybrid reasoning layer described in Section 2.5 performs contextual hazard assessment using the current scene graph and a rolling temporal graph spanning the most recent 10 s. While this short-horizon representation is sufficient for detecting immediate motion-based risks such as approach behaviour, proximity escalation, and short-term anomaly persistence, many safety-critical patterns in underground mining evolve over longer periods and may be difficult to detect within the bounds of a short rolling window alone. Examples include repeated unsafe interactions between the same personnel and equipment, gradual increases in uncertainty near structural elements, recurrent occupancy of blind zones, and repeated emergence of similar anomalous configurations at specific spatial locations. Such patterns may appear weak or ambiguous within a short temporal window but become informative when interpreted relative to historical graph states. This motivates the use of a Graph Retrieval-Augmented Generation (GraphRAG) [51] based longitudinal reasoning pipeline, as illustrated in Figure 8.

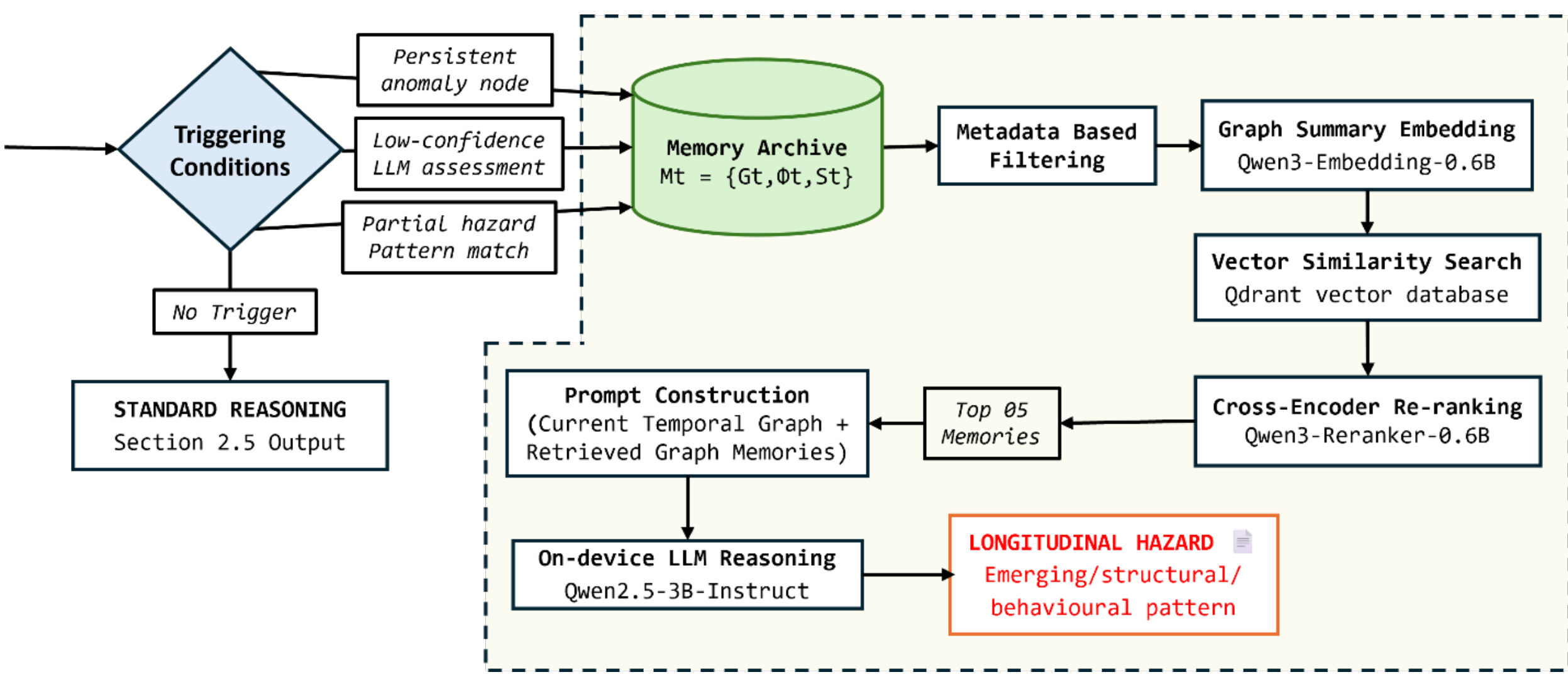

*Figure 8: GraphRAG-based longitudinal hazard reasoning pipeline*

The proposed GraphRAG pipeline maintains a persistent archive of historical graph states, retrieves the most relevant prior observations when longitudinal reasoning is required, and injects these retrieved memories into the language model prompt together with the current graph representation. Consequently, the language model receives both the current scene configuration and selected historical evidence, enabling it to determine whether the current situation represents an isolated event, a recurring pattern, or an escalating safety condition.

### 2.6.1 Persistent Scene Graph Memory Unit

To support historical reasoning, the system maintains a persistent archive of previously observed graph states. At each reasoning step, a compact memory unit is generated from the current temporal graph and stored in the memory archive. The memory representation at time $\boldsymbol{t}$ is defined as:

$$M_t = \{G_t, \Phi_t, S_t\}$$

The graph component $G_t$ corresponds to the relational structure described in Section 2.3, preserving object identities and spatial relations. The attribute set $\Phi_t$ stores contextual information derived from the temporal graph and reasoning layers, including motion descriptors of tracked objects, deterministic safety flags generated by the rule-based evaluation stage, anomaly persistence indicators derived from predictive entropy, and prior risk interpretations produced by the language model. By combining the structural graph representation with these higher-level descriptors, each memory unit captures not only the spatial configuration of the environment but also the contextual interpretation assigned to that configuration during earlier reasoning steps. $S_t$ is defined as a structured textual summary of $G_t$, encoding the dominant object classes, key relational predicates, motion states of tracked entities, active safety flags, and previously inferred risk information. This compact representation enables the graph state to be embedded and indexed for efficient similarity search without requiring the full graph structure to be processed during retrieval.

### 2.6.2 Memory Retrieval and Ranking

Historical retrieval is not performed at every observation step. The deterministic safety evaluation in Section 2.5 handles explicit rule violations with negligible latency, while the contextual reasoning module interprets most hazards using the current 10 s temporal graph. Retrieval from the memory archive is activated only when additional historical context is needed. In this study, three operational triggers are used: an anomaly node persists for more than 3 s, equivalent to 12 snapshots within the current temporal window; the contextual reasoning module returns a hazard status of developing or unresolved; or the current scene graph embedding has a cosine similarity greater than 0.70 with at least one archived record in the Qdrant store. These triggers indicate that the current observation may represent a persistent anomaly, an uncertain relational hazard, or a recurring safety pattern. In such scenarios, the system queries the historical memory archive $\mathcal{M} = \{M_1, M_2, \dots, M_{t-1}\}$.

The GraphRAG retrieval pipeline is implemented using the LlamaIndex framework [52] with a locally deployed Qdrant vector database [53]. Qdrant is selected for its support for efficient vector similarity search combined with structured metadata filtering, enabling retrieval based on both semantic similarity and operational constraints. At each reasoning step, the graph summary $S_t$ is first converted into a dense vector representation using the Qwen3-Embedding-0.6B model [54]. This model is selected for its strong performance on multilingual embedding benchmarks (≈ 64 MTEB score), while maintaining a compact 0.6B parameter count and low reported standalone inference latency (20–22 ms per query), making it suitable for on-device deployment.

The embedding of $S_t$, together with the associated metadata $\Phi_t$, is stored in the Qdrant index. The metadata includes object classes, relational interaction types, motion states, anomaly indicators, and deterministic safety flags derived from the scene and temporal graphs. During retrieval, the embedding of the current graph summary is used to query the Qdrant index. The database performs vector similarity search to identify historical graph memories that are semantically similar to the current scene configuration while simultaneously applying metadata constraints to restrict candidate memories to operationally compatible graph states. This retrieval stage produces a ranked list of candidate graph summaries based on embedding similarity. To further improve retrieval precision, the candidate memories are refined using the Qwen3-Reranker-0.6B model, which is part of the same embedding series designed specifically for ranking tasks in retrieval pipelines [54]. The reranker operates as a cross-encoder that jointly evaluates the current graph summary and each retrieved candidate summary. This produces refined relevance scores that prioritise candidates whose relational structure and contextual attributes most closely match the current scene configuration.

The top five highest-ranked graph memories are selected and provided to the reasoning module as historical context [55]. The final contextual analysis is performed using the Qwen2.5-3B-Instruct language model operating directly on the monitoring device. The reasoning prompt contains the serialised representation of the current temporal graph together with the retrieved historical graph summaries. This combined context allows the language model to interpret the current situation within the broader operational history of the monitored environment and identify longitudinal hazard patterns that cannot be inferred from the short temporal window alone.

### 2.6.3 Longitudinal Hazard Interpretation

Using this combined context, the reasoning module analyses relational, kinematic, and anomaly-level similarities between the current graph and retrieved memories to identify longitudinal hazard patterns that cannot be detected within a single temporal window. Because each memory unit $\Phi_t$ preserves prior reasoning outputs alongside kinematic and anomaly descriptors, the system can also analyse how previously identified safety conditions evolve over time. This allows the reasoning module to determine whether an observed situation represents a transient occurrence, a recurring unsafe interaction pattern, or an escalating hazard condition, thereby supporting proactive safety intervention before critical thresholds are reached.

## 3. Results

### 3.1 Semantic Detection Performance

The proposed sparse 3D Minkowski UNet was trained using the two-stage strategy described in the methodology, consisting of self-supervised contrastive pretraining on the designated pretraining pool followed by supervised fine-tuning on the annotated training subset. The annotated subset spans three distinct acquisition configurations: fixed multi-sensor simulated longwall scans, mobile single-sensor simulated longwall traversals, and fully annotated real underground roadway point clouds. This diverse training set was designed to expose the model to both the geometric consistency of synthetic environments, and the authentic structural variation, sensor noise, and environmental irregularities present in real field data. Training was performed on a workstation equipped with 64 GB of RAM and an Intel Core i9-13900K CPU, and an NVIDIA GeForce RTX 4080 GPU with 16 GB of VRAM.

During supervised fine-tuning, the contrastive-initialised model started with a substantially lower cross-entropy loss than the model trained from random initialisation, indicating that the backbone had already learned useful structural features before supervised learning began. This advantage was most pronounced during the first 20 epochs, when the contrastive-pretrained model reduced both training and validation loss more rapidly. After this initial reduction, the losses for both models continued to improve gradually, as shown in Figure 9.

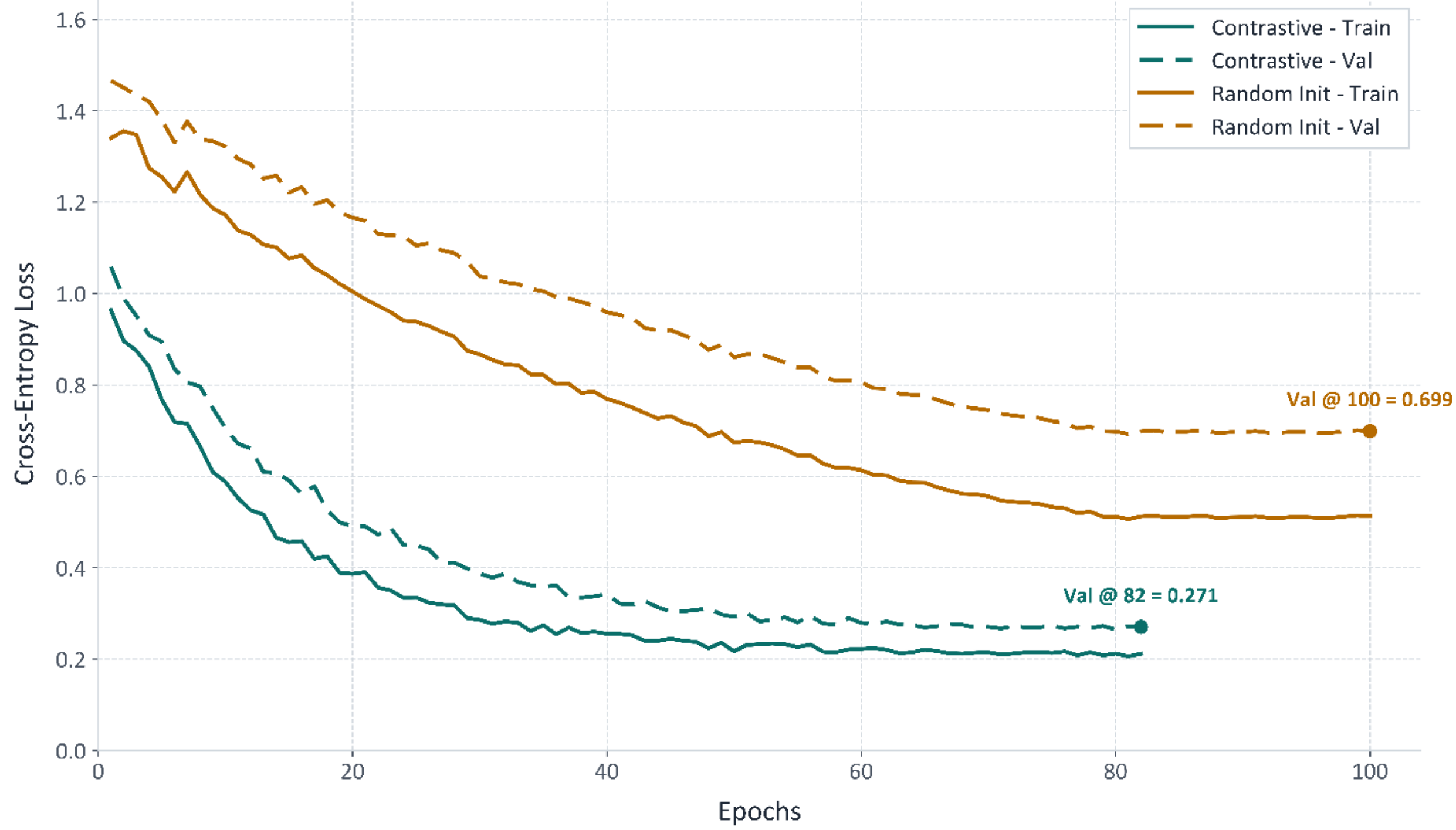

*Figure 9: Training and validation loss curves comparing contrastive-pretrained initialisation versus random initialisation across supervised fine-tuning epochs. The contrastive-pretrained model exhibits faster convergence and lower final validation loss.*

The validation curves further demonstrated the benefit of contrastive initialisation in the limited labelled-data regime. The contrastive-pretrained model reached a minimum validation loss of 0.271 at epoch 82, whereas the randomly initialised model improved more slowly and reached a substantially higher validation loss of 0.699 at epoch 100. This difference demonstrated that contrastive pretraining improved optimisation efficiency and generalisation performance during supervised fine-tuning.

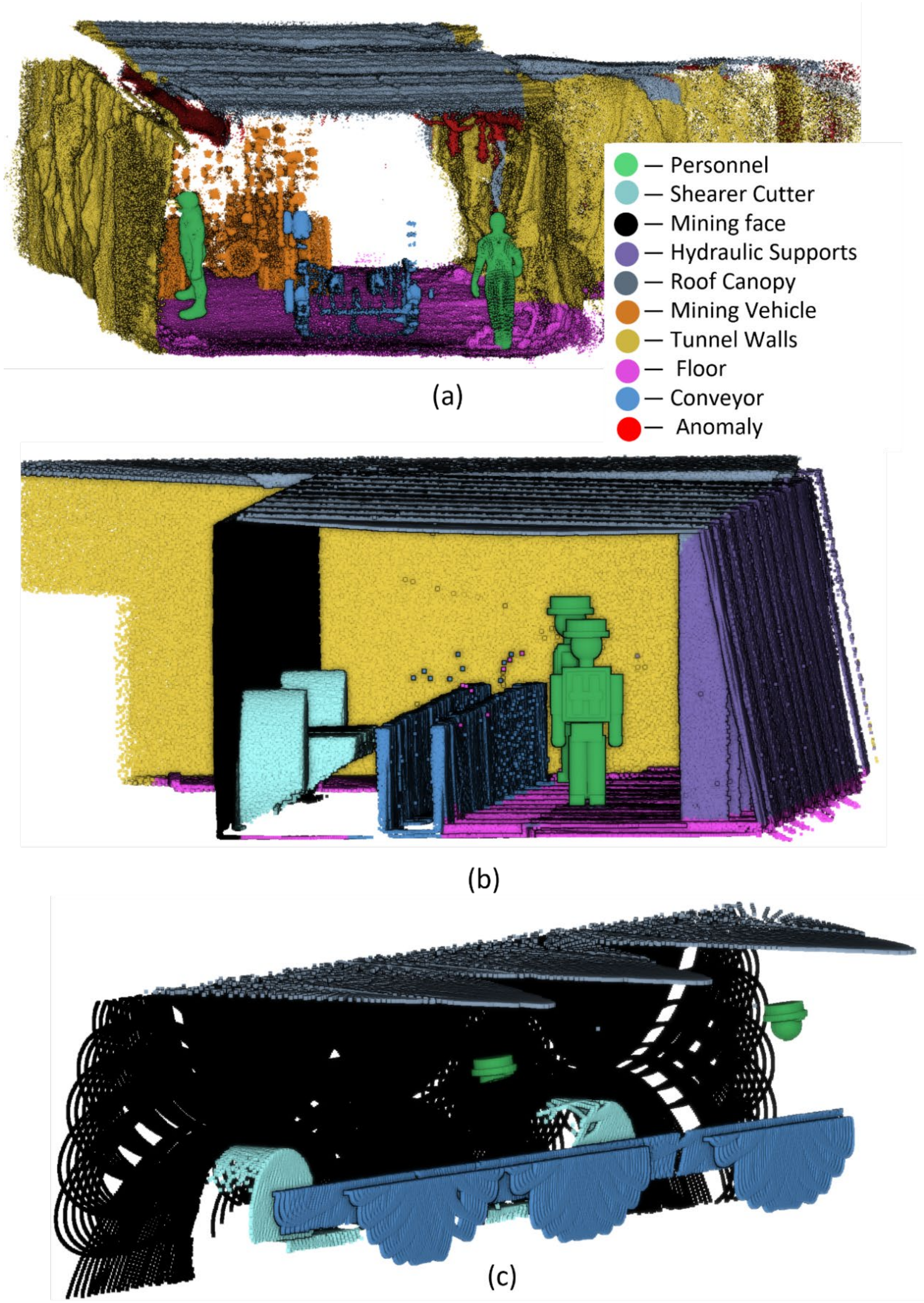

*Figure 10: Qualitative segmentation results across data domains: (a) real underground roadway scan with inserted object instances; (b) simulated roadway scene reconstructed from a mobile sensing unit; (c) simulated longwall scene from fixed-mounted sensing units. Colour coding follows the class legend shown at right.*

The qualitative segmentation outputs illustrated in Figure 10 provide further insight into detection behaviour across the two data domains. In the simulated scenes (middle and bottom panels), object boundaries are consistently sharp and well defined, reflecting the geometric regularity of the synthetic environments. Structural elements such as hydraulic supports, the roof canopy, conveyor, and tunnel walls are segmented with clean, spatially coherent boundaries, while dynamic objects, including the shearer cutter and personnel, are accurately delineated with minimal label leakage into adjacent classes. This boundary precision is partly attributable to the idealised surface geometry and controlled point density in the simulations. In contrast, segmentation results on the real underground roadway data (top panel) show a visually noisier appearance consistent with field-collected point clouds. Boundaries associated with the roof and wall surfaces are less geometrically regular; however, key object classes remain well separated, particularly personnel and mining vehicles. Despite real-world surface irregularities, variable point density, and occlusion effects, all major semantic classes are correctly identified and spatially localised, demonstrating that the model generalises effectively from synthetic training data to authentic underground conditions. Anomaly regions, shown in red, are localised to structurally irregular areas along the roof surface, consistent with roof bolts and other protrusions that fall outside the predefined semantic classes. Overall, these qualitative results indicate consistent segmentation behaviour across both synthetic and real underground domains, which is examined quantitatively below.

To further confirm these observations, segmentation performance was evaluated using voxel-level metrics, including overall accuracy, mean Intersection over Union (mIoU), and per-class IoU. Across all semantic categories, the model achieved an Overall Accuracy (OA) exceeding 92%, indicating reliable classification of both structural and dynamic elements in the underground environment. For a more balanced assessment across classes with different spatial extents, performance was further analysed using Intersection over Union (IoU). The resulting per-class IoU values demonstrate stable segmentation performance across the nine semantic categories, with scores of 0.86 for Personnel, 0.95 for Shearer, 0.91 for Hydraulic Supports, 0.86 for Conveyor, 0.81 for Roof Canopy, 0.86 for the Mining Face, 0.87 for Mining Vehicle, 0.81 for Walls, and 0.84 for Floor, corresponding to an overall mIoU of 0.86. While structural classes generally achieve slightly higher IoU, the lower scores for Walls and Floor likely reflect boundary irregularities and sensing variability present in the real underground data.

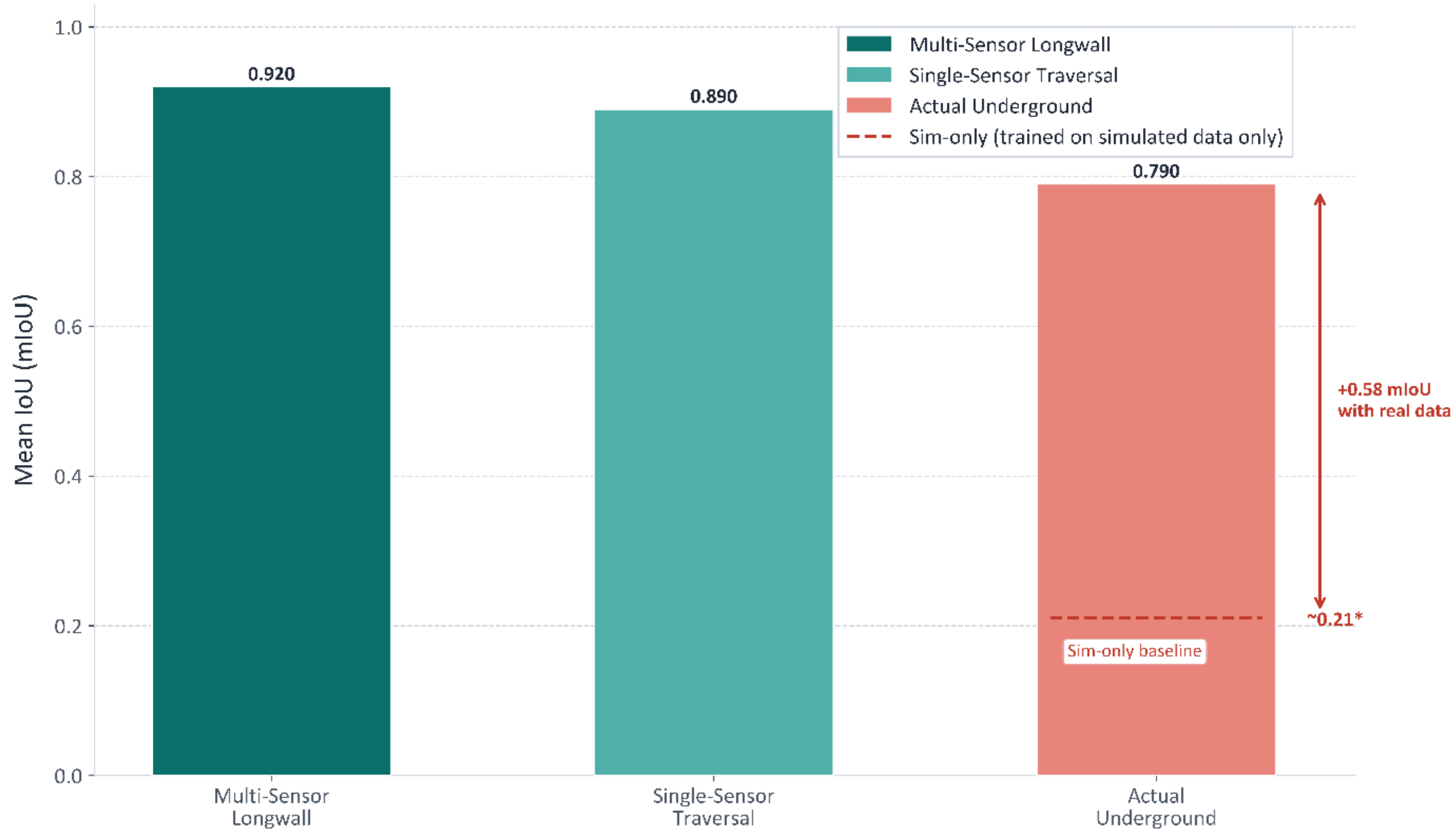

*Figure 11: Mean IoU across acquisition configurations*

Beyond class-level segmentation accuracy, another objective of the evaluation was to analyse the model's robustness across different data acquisition configuration using the held-out testing set. As shown in Figure 11, the model achieves an mIoU of 0.92 on multi-sensor simulated longwall scans and 0.89 on simulated single-sensor traversal data, while maintaining an mIoU of 0.79 on real underground point clouds. Although performance decreases when transitioning from simulated to real environments, the model retains useful segmentation capability under real operational conditions. To further illustrate the influence of domain differences, an additional experiment was conducted where the model was trained only on simulated data and evaluated directly on real underground scans. Under this configuration, segmentation performance drops significantly to approximately 0.21 mIoU, indicating the substantial gap between simulated sensing conditions and real underground environments. As illustrated in Figure 11, incorporating real underground samples during training improves real-world performance from ~0.21 mIoU to 0.79 mIoU, demonstrating the importance of including representative real data to bridge the simulation-to-reality gap.

*Table 2: Comparative segmentation performance of the proposed model against PointNet++ and Point Transformer V3*

| Model | Training Strategy | Overall Accuracy OA (%) | Mean Intersection over Union (mIoU) | FPS | GPU Memory (MB) |
|---|---|---|---|---|---|
| PointNet++[14] | Supervised Learning | 84.2 | 0.55 | 4.5 | 461 |
| Point Transformer V3 [56] | Supervised Learning | 86.4 | 0.58 | 17.4 | 301 |

| Proposed Model | Supervised Learning | 87.4 | 0.62 | 30.2 | 94 |
|---|---|---|---|---|---|
| Proposed Model | Contrastive Pretraining + Fine-Tuning | 92.7 | 0.86 | 30.2 | 94 |

To contextualise the performance of the proposed approach, the segmentation model was evaluated against PointNet++ and Point Transformer V3 under identical conditions, including the same data splits, preprocessing, input features, augmentation settings, and training settings, as summarised in Table 2. These models provide two useful comparison points: PointNet++ as a widely used point-cloud segmentation baseline, and Point Transformer V3 as a more recent transformer-based method for 3D perception. PointNet++ achieved 84.2% OA and 0.55 mIoU, with an inference speed of 4.5 FPS and 461 MB GPU memory usage. Point Transformer V3 improved the result to 86.4% OA and 0.58 mIoU, while increasing the inference speed to 17.4 FPS and reducing memory usage to 301 MB. In comparison, the proposed model trained with supervised learning alone achieved 87.4% OA and 0.62 mIoU, while operating at 30.2 FPS with only 94 MB of GPU memory.

The proposed model improved further when contrastive pretraining was applied prior to supervised fine-tuning, achieving 92.7% OA and 0.86 mIoU. This result shows that the performance gain is not only due to the sparse architecture, but also to the stronger feature representation learned from unlabelled underground point clouds during contrastive pretraining. Although the improvement in OA is moderate compared with the larger gain in mIoU, this is expected in a class-imbalanced dataset where OA is strongly influenced by large structural classes such as walls, floor, and roof canopy. In contrast, mIoU gives a more balanced view across all classes and is more sensitive to improvements in smaller safety-critical objects such as personnel, shearer, conveyor, and mining vehicles. Therefore, the substantial mIoU improvement indicates that the proposed model provides more reliable segmentation of objects important for downstream hazard reasoning. Overall, compared with PointNet++ and Point Transformer V3, the proposed model achieves higher segmentation accuracy, faster inference, and substantially lower GPU memory demand, demonstrating its suitability for real-time underground edge deployment.

### 3.2 Evaluation of Graph-Based Safety Reasoning and Hazard Detection

Evaluating the proposed hybrid cognitive reasoning pipeline on isolated hazard instances does not reflect real-world conditions in underground mining. In practice, most observations correspond to normal safe operation, with hazardous events occurring only intermittently. Evaluating only pre-selected hazard frames can therefore bias performance by inflating detection rates while obscuring false positives during routine operation. To address this, the evaluation was conducted on continuous, unsegmented data streams across three environments: the simulated longwall panel, the simulated tunnel roadway environment, and the real underground roadway dataset. Approximately 95% of graph snapshots represented normal safe operation, while the remaining 5% contained labelled hazard events. In total, 115 labelled hazard instances were evaluated, covering deterministic threshold violations, interaction-based hazards, structural and environmental anomalies, and temporal hazard patterns, as summarised in Table 3.

*Table 3: Summary of hazard scenarios used for evaluation, including scenario counts and simulation method*

| Hazard Scenario | Count | Simulation Method |
| --- | --- | --- |
| Personnel within 1.5 m of active mining equipment | 20 | Personnel trajectories were generated to pass active equipment at controlled separations (0.5 m, 1.0 m, 1.5 m) under varying relative-motion conditions. |
| Personnel and equipment on a converging trajectory with varying TTC thresholds | 20 | Personnel and equipment assigned controlled initial separations and motion vectors to produce predefined TTC values. |
| Personnel in the equipment blind spot, occluded by supports or structure | 10 | Personnel and equipment followed trajectories with different viewing directions, resulting in blind-spot encounters during equipment approach or operation. |
| Multiple personnel in a congested proximity zone near active equipment | 10 | Two or three personnel placed simultaneously near active equipment with varied motion states. |
| Unexpected obstacle or unmodelled structure near personnel or the operational path | 10 | Additional obstacle objects or unmodelled structural features inserted near personnel locations or machinery travel paths. |
| Personnel proximity pattern recurring across consecutive frames | 10 | Personnel repeatedly approached active equipment across consecutive frames while remaining below explicit proximity violation thresholds. |
| Progressive structural deterioration at a fixed location across frames | 5 | A fixed structural surface was incrementally deformed across consecutive frames to simulate progressive deterioration. |
| Personnel dwell time increasing progressively within a high-risk zone | 5 | Personnel remained within a hazardous zone for progressively longer durations across consecutive frames. |
| Structural anomaly recurring at the same spatial location across multiple sessions | 5 | Similar anomaly instances were reintroduced at the same spatial location across multiple simulation runs. |
| Equipment behaviour deviating from nominal operational patterns | 5 | Equipment motion deviated from nominal operation using irregular speed, pauses, and directional changes across consecutive frames. |
| Near-miss interaction pattern recurring across consecutive frames | 5 | Personnel repeatedly entered blind-spot or near-miss regions across consecutive temporal windows |
| Personnel prolonged no-motion anomalies | 5 | Personnel remained motionless within operational zones across consecutive frames for prolonged durations. |

| Personnel and equipment operating under reduced visibility conditions | 5 | Light source intensity changed to achieve different visibility conditions |
|---|---|---|
| Total | **115** | |

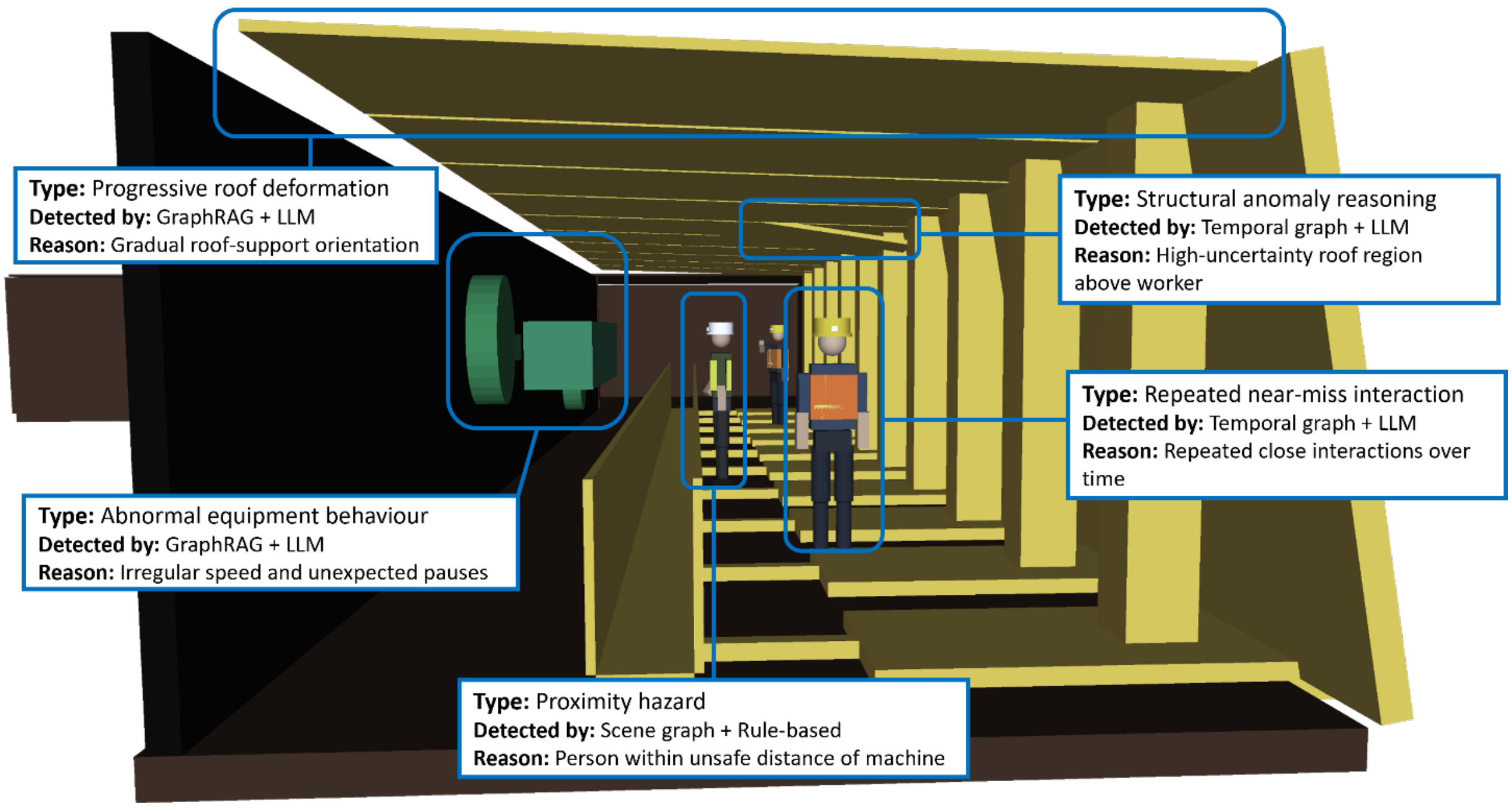

*Figure 12: Hazard scenarios embedded in continuous ROS Gazebo scene streams and used for evaluation of the proposed reasoning pipeline. The hazards shown are illustrative examples and are not all injected simultaneously in a single scenario, while their associated reasoning pathways are described in the following section.*

Hazard scenarios were generated using two complementary evaluation settings. In the first setting, controlled hazards were embedded within continuous ROS Gazebo scene streams using the configurations defined in Table 3. Personnel agents, mining equipment, and structural elements were programmatically scripted through controlled trajectories, motion behaviours, separations, and interaction conditions to generate the spatial, contextual, and temporal hazards considered in the evaluation. Representative examples are illustrated in Figure 12. Within the simulator, virtual sensing devices generated real-time scene data, which were continuously streamed through the proposed hybrid cognitive reasoning pipeline, mirroring the operational data flow of the deployed system. This enabled hazard assessment under continuous operational conditions rather than isolated injected events, while providing a precise and repeatable mechanism for evaluating diverse hazard conditions across both the longwall and tunnel environments.

In the second evaluation setting, hazards were constructed within reconstructed real underground roadway point cloud streams. Previously scanned point clouds of personnel, mining vehicles, and conveyor components were registered into roadway sections at controlled spatial configurations corresponding to the scenarios defined in Table 3. This preserves the authentic geometry, sensor noise, occlusion effects, and structural characteristics of the real environment while enabling controlled hazard construction under realistic sensing conditions. Representative examples are shown in Figure 13. The resulting point cloud streams were provided directly as input to the proposed hybrid cognitive reasoning pipeline, allowing the same detection and reasoning framework used in simulation to be evaluated on real underground data.

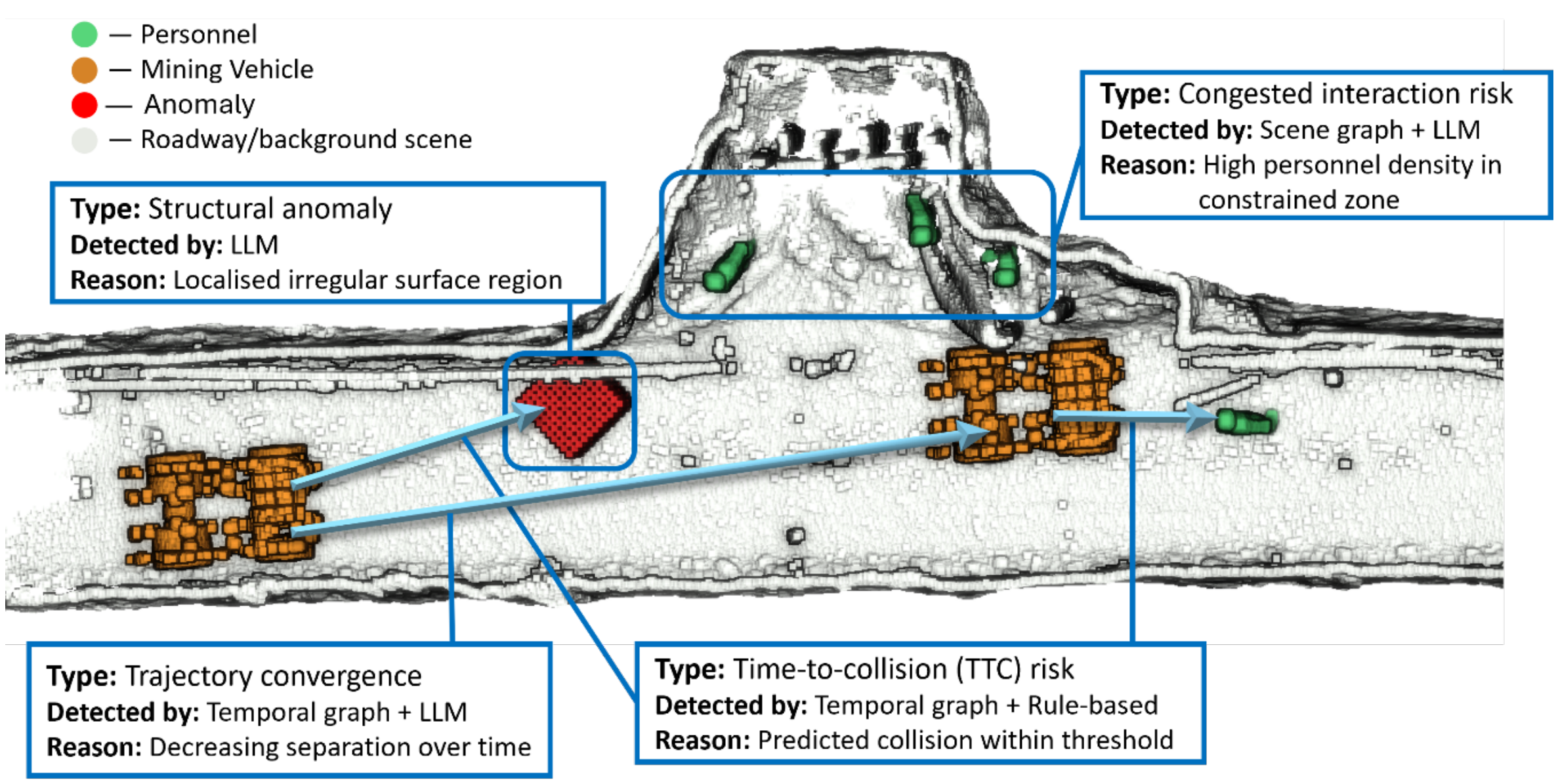

*Figure 13: Representative hazard instances constructed within reconstructed underground point cloud streams for the evaluation of the proposed reasoning pipeline. The hazards shown are illustrative examples only and are not all introduced simultaneously in a single scenario.*

Each evaluation session consisted of two phases. The first represented nominal safe operation and provided a continuous baseline for measuring false positives. In the second phase, hazard conditions from Table 3 were introduced at randomised intervals within the same continuous stream. Hazards were introduced one at a time to isolate the system response to each event and to measure detection latency without overlap between concurrent hazards. Ground-truth hazard onset and resolution times were recorded programmatically for latency measurement.

### 3.2.1 Deterministic Rule-Based Pre-Evaluation

The deterministic rule-based layer was evaluated first to establish a baseline and to characterise the limits of threshold-based safety enforcement. This layer is designed to operate as a near-zero-latency safety gate that evaluates multiple explicit hazard conditions directly from the static and temporal scene graphs and associated point cloud representations. These include rule-based checks for unsafe personnel–equipment proximity, trajectory convergence indicative of collision risk, blind-spot occupancy inferred from relative geometry and orientation, congested multi-person proximity interactions, and reduced visibility conditions derived from the colourised point cloud. Alerts are triggered whenever these predefined geometric, kinematic, or sensing constraints are violated.

Across the full set of 115 evaluation scenarios, the deterministic rule engine correctly detected all 65 scenarios within its rule-defined scope. No false positives were observed during baseline operation. Detection within scope was consistent and complete, with no missed detections for any applicable scenario. As shown in Figure 14, rule-engine execution remained in the sub-millisecond range under typical scene complexity, with median latency around 0.7–0.8 ms for scenes containing approximately 10 objects, which represents a common interaction scale in the evaluated environments. As object count and pairwise interactions increased, execution time rose gradually, reaching approximately 1.5 ms even for scenes exceeding 3000 pairwise checks with more than 100 objects. The component-level timing breakdown further shows that visibility assessment accounts for the largest share of computation, whereas proximity, congestion, blind-spot, and trajectory checks contribute only minor additional overhead. Although visibility evaluation is comparatively more expensive due to rolling temporal analysis over colourised voxels, the overall rule-check latency remains sufficiently low to support a near-zero-latency safety gate.

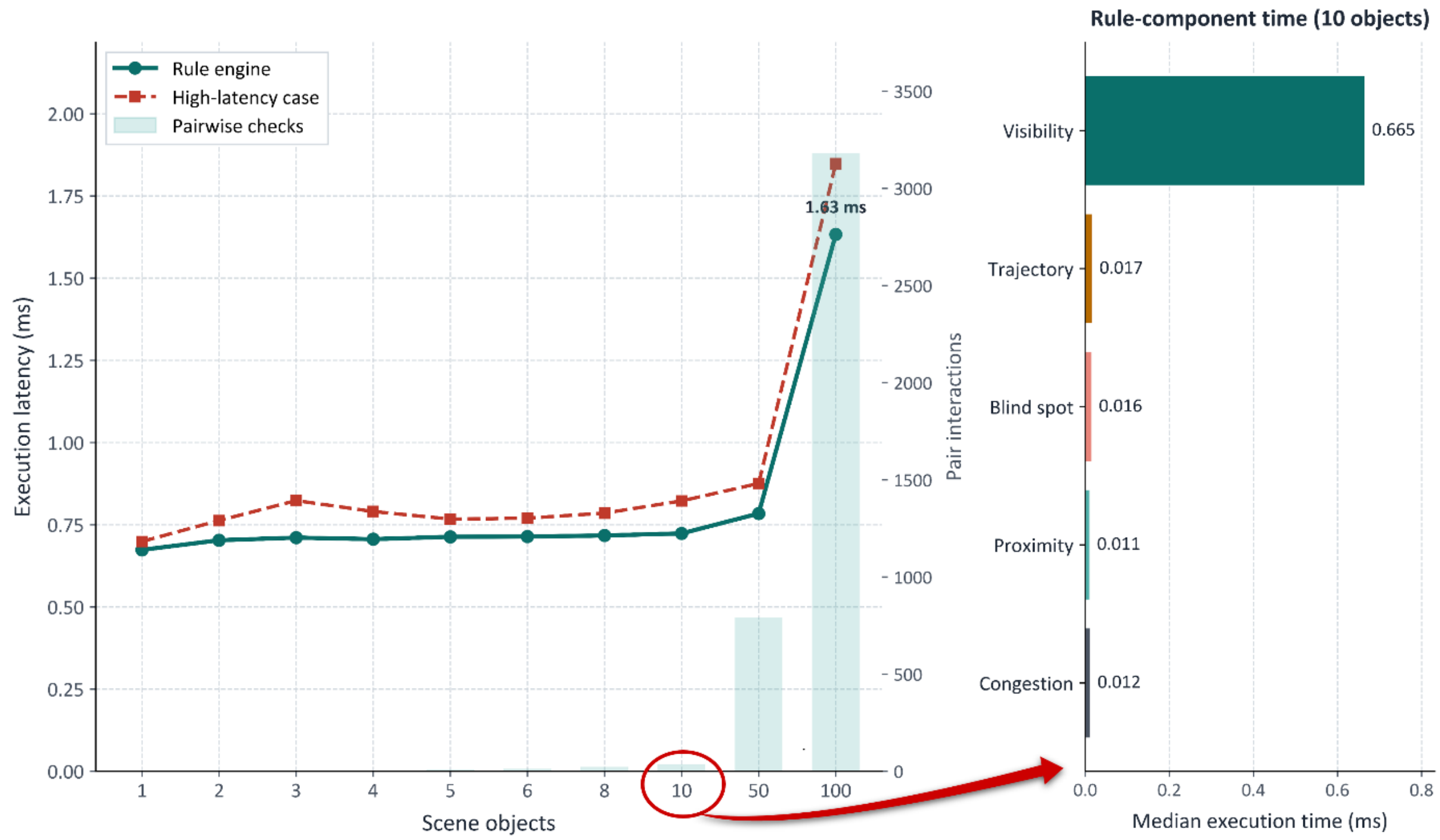

*Figure 14: Computational latency of the deterministic rule engine under increasing interaction complexity and breakdown of median execution time across rule components.*

The remaining 50 scenarios, representing 43% of the evaluation set, involved hazards requiring temporal persistence, contextual interpretation, anomaly reasoning, or behavioural pattern assessment, which fall outside the capability of threshold-based evaluation. These scenarios require contextual interpretation or temporal reasoning beyond instantaneous measurements. This result highlights that while the deterministic layer provides reliable and immediate enforcement of strict safety constraints, it cannot capture the full range of underground hazard conditions. This limitation motivates the need for the contextual reasoning and longitudinal analysis layers presented in the following sections.

### 3.2.2 LLM-Based Reasoning Evaluation

To address the coverage limitations of the deterministic rule layer, candidate language models were first screened based on deployment feasibility, including context capacity, inference throughput, and hardware compatibility. The evaluation was conducted on a local workstation equipped with an NVIDIA GeForce RTX 4080 GPU (16 GB VRAM) to reflect on-device deployment constraints. As shown in Table 4, several models were excluded due to insufficient context window, excessive latency, or deployment constraints. Based on this evaluation, three models—DeepSeek-R1-Distill-Qwen-1.5B, Phi-4-mini-instruct, and Qwen2.5-3B-Instruct—were shortlisted for further analysis.

*Table 4: Hardware feasibility benchmark for candidate model screening (Shortlisted models are highlighted in green)*

| Model | Params (B) | Context Window | Weights Size (GB) | Load Time (s) | Tokens per second | Decision |
|---|---|---|---|---|---|---|
| DeepSeek-R1-Distill-Qwen-1.5B | 1.78 | 131,072 | 3.55 | 6.5 | 36.3 | Shortlisted |
| Phi-4-mini-instruct | 3.84 | 262,144 | 7.67 | 6.2 | 29.8 | Shortlisted |
| Qwen2.5-3B-Instruct | 3.09 | 131,072 | 6.17 | 5.3 | 30.8 | Shortlisted |
| Phi-3-mini-4k-instruct | 3.82 | 4,096 | 7.64 | 5.3 | 37.1 | Insufficient context window |
| Gemma-3-4b-it | 4.30 | 128,000 | 8.60 | 7.7 | 17.2 | Low throughput and high memory usage |
| Qwen2.5-7B-Instruct | 7.62 | 131,072 | 15.23 | 14.2 | 0.31 | Infeasible latency |
| Llama-3.2-3B-Instruct, Nemotron-H-4B-Instruct | — | — | — | — | — | Deployment constraints |

The shortlisted LLMs were evaluated within the hybrid hazard detection pipeline, using the serialised temporal scene graph as input, which comprises semantic, geometric, motion, uncertainty, and deterministic safety attributes, as described in Section 2.5.2. Deterministic rule outputs were treated as established signals within the prompt, while the LLM provided additional contextual interpretation for hazards not captured by fixed thresholds, without altering rule-based detections. Two evaluation metrics were recorded for each model: reasoning accuracy, defined as the proportion of scenarios in which the model correctly identified hazard conditions and produced an appropriate risk assessment; and structured output stability, defined as the proportion of responses that fully adhered to the predefined output schema, without formatting failures or malformed fields. The comparative performance of the shortlisted models is summarised in Table 5.

*Table 5: LLM reasoning performance on non-rule-based scenarios*

| Model | Structured Outputs (Valid / 25) | Reasoning Accuracy (Correct / 25) |
|---|---|---|
| Phi-4-mini-instruct | 21 (84%) | 16 (64%) |
| DeepSeek-R1-Distill-Qwen-1.5B | 23 (92%) | 19 (76%) |

| Qwen2.5-3B-Instruct | 23 (92%) | 22 (88%) |
|---|---|---|

Of the 50 hazards outside the scope of deterministic rules, 25 involved contextual conditions that could be assessed from the current scene and temporal graph without historical retrieval. These scenarios were used to benchmark standalone LLM reasoning performance. As summarised in Table 5, all three models produced mostly valid structured outputs, but their reasoning accuracy varied across the contextual hazard scenarios. Qwen2.5-3B-Instruct achieved the best overall balance between reasoning accuracy and output stability and was therefore selected as the reasoning model for subsequent longitudinal reasoning evaluations. Using this selected model, standalone LLM reasoning increased hazard coverage from 65 to 87 out of 115 scenarios, improving overall coverage from 57% to 76%. This demonstrates that the LLM layer extends detection beyond threshold-based hazards by resolving contextual conditions from the current scene and temporal graph. The remaining unresolved scenarios require longer-term historical context and are addressed in the following section.

Beyond quantitative detection performance, qualitative analysis revealed an unexpected reasoning behaviour not explicitly targeted by the designed hazard scenarios. In real underground roadway scenes, the semantic segmentation model consistently produced high-entropy regions along the roof surface, corresponding to the irregular geometry of roof bolts installed through the roof canopy into the surrounding rock mass to support the roadway. As these protruding structures were not included as a labelled semantic class, they appeared as clustered nodes in the scene graph. Both DeepSeek-R1-Distill-Qwen-1.5B and Qwen2.5-3B-Instruct correctly interpreted these clusters as roof support infrastructure based on their spatial distribution, geometry, and alignment along the roof (Figure 15). Qwen2.5-3B-Instruct further extended this reasoning by analysing their spatial distribution across the roadway and identifying a localised region of reduced inferred bolt density, suggesting a potential structural support deficiency. Notably, this hazard was not part of the predefined evaluation scenarios and was inferred directly from spatial evidence. These results demonstrate a broader capability beyond rule-based detection, showing that low-level anomaly cues can be synthesised into interpretable, context-aware safety insights without explicit supervision or prompt instruction.

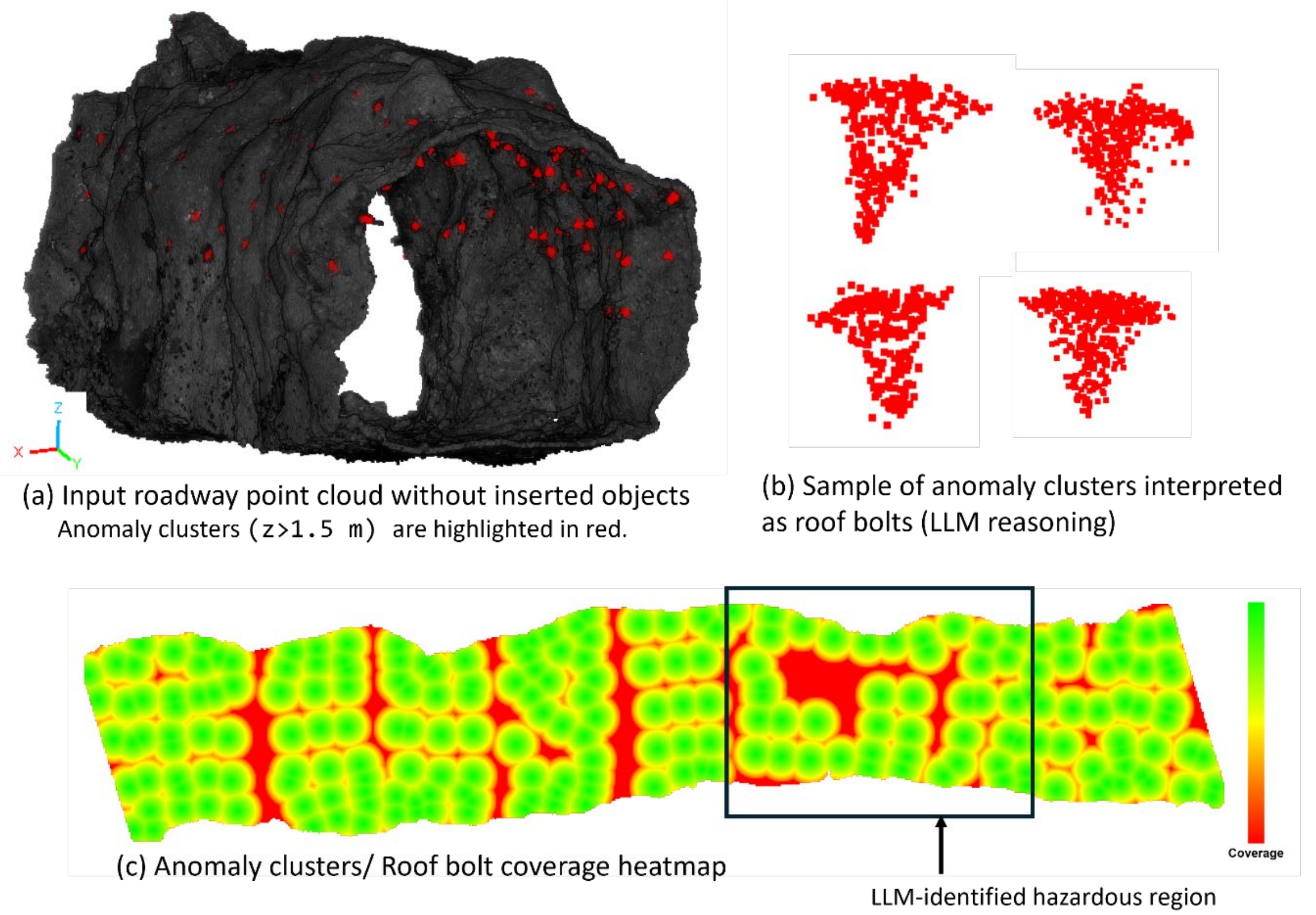

*Figure 15: LLM-driven interpretation of anomaly nodes for hazard detection (a) Input roadway point cloud (internal objects removed for clarity) with detected high-entropy anomaly nodes (red); (b) Representative anomaly clusters extracted from the scene; (c) Roof bolt coverage heatmap inferred by Qwen2.5-3B-Instruct from anomaly nodes, with the highlighted region indicating reduced bolt density and a potential structural support deficiency.*

### 3.2.3 Direct Point-Cloud LLM Feasibility Validation

Before finalising the structured scene-graph interface, a feasibility test was conducted to examine whether a native point-cloud language model could bypass the proposed perception and graph-construction stages. At the time of evaluation, no practical mainstream commercial multimodal LLM was identified that could locally process full scene xyzrgb point clouds under the required hardware constraints. Therefore, PointLLM-7B-v1.2 was selected as a locally runnable research baseline for direct point cloud reasoning [36].

The test was performed on the same workstation used for the main model evaluation, equipped with an NVIDIA RTX 4080 GPU with 16 GB VRAM under WSL CUDA. The average point cloud size in the proposed dataset is approximately 2.7 million points, while PointLLM operates on an 8192 point xyzrgb input. To make the test as favourable as possible, two input settings were evaluated. First, the model was given the colourised mine point cloud directly. Second, the model was given a semantically guided version of the same scene, where the Minkowski UNet segmentation output was encoded using class-based colors, and the prompt included the corresponding class legend. In this second setting, the model received both the point cloud and explicit semantic guidance during inference. PointLLM was computationally feasible but close to the practical memory limit of the workstation. The measured peak GPU memory usage was approximately 13.1 to 13.2 GB, indicating that a 16 GB GPU is a practical minimum for this configuration. After loading, response generation required approximately 11 to 13 s, with a first token latency of approximately 8.8 s.

Across both input settings, PointLLM accepted the sampled xyzrgb input and generated natural-language descriptions, but its outputs were not sufficiently reliable for safety-critical mine-scene interpretation. The model did not consistently follow the requested structured JSON format, hallucinated absent classes such as humans despite the expected human count being zero, and produced inconsistent descriptions of mine-specific classes and spatial relationships. In contrast, the proposed segmentation and scene graph pipeline produced a deterministic semantic summary of the scene, correctly identifying the relevant object classes, confirming the absence of personnel, and extracting geometry-based spatial relations between nearby objects. These results support the use of a structured scene graph interface rather than direct point cloud prompting. The following section evaluates how this structured representation can be extended with GraphRAG-based temporal history for longitudinal hazard reasoning.

#### 3.2.4 Reasoning over Temporal History via GraphRAG

Following the selection of Qwen2.5-3B-Instruct as the on-device reasoning model in Section 3.2.2, the complete perception-to-reasoning framework, including GraphRAG-based historical context retrieval, was evaluated across all 115 hazard scenarios. The deterministic rule layer correctly identified 65 scenarios, standalone LLM reasoning identified a further 22 contextual hazards from the current scene and temporal graph, and GraphRAG-augmented longitudinal reasoning correctly identified 20 of the remaining 25 scenarios involving recurring interactions, persistent behavioural deviations, repeated structural anomalies, and progressive deterioration patterns that cannot be interpreted from the current observation window alone. Combined, these three layers achieved overall coverage of 107 out of 115 hazard scenarios, corresponding to approximately 93% hazard coverage, demonstrating that longitudinal memory substantially extends detection beyond both threshold-based rules and standalone contextual reasoning.

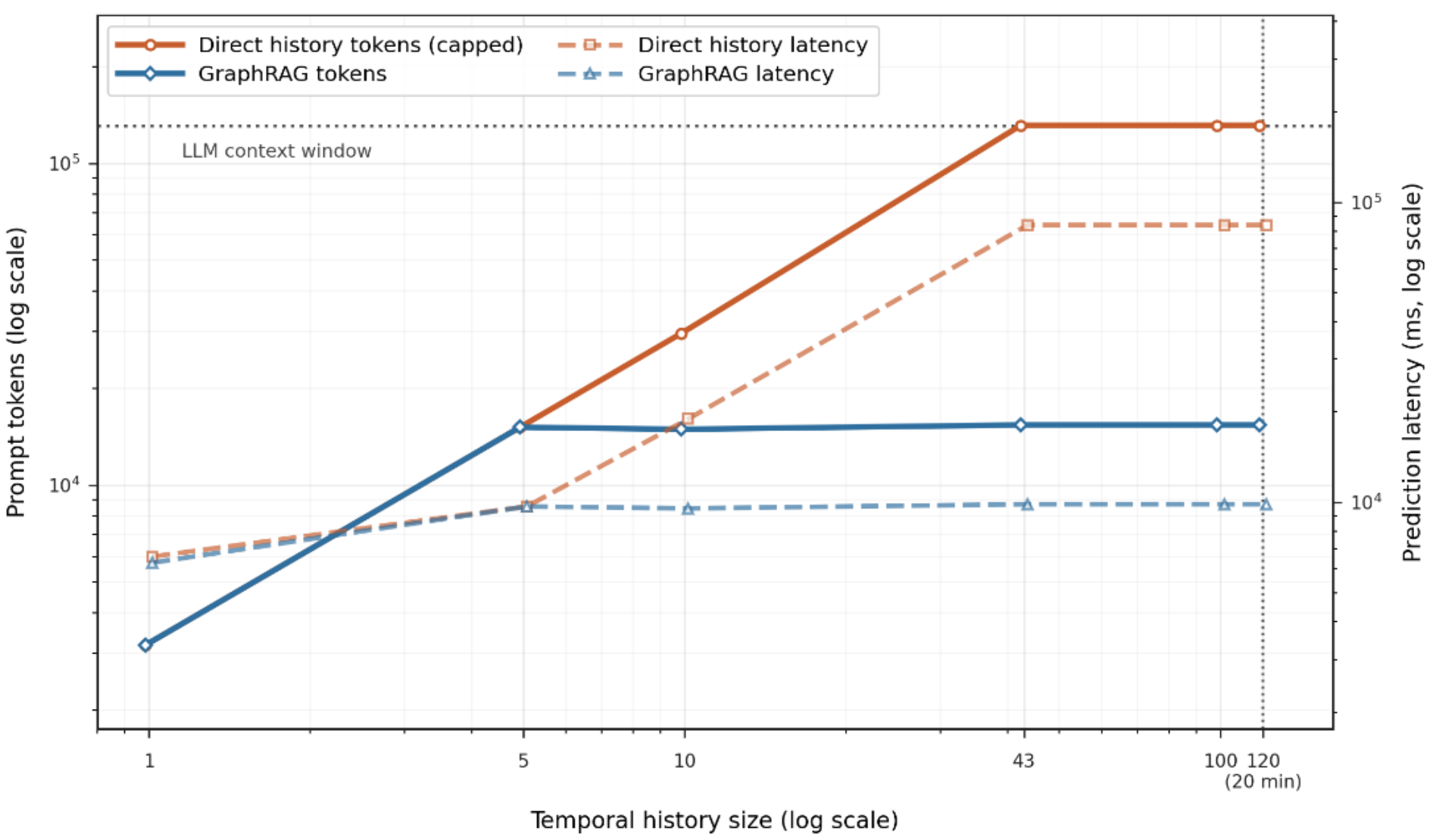

*Figure 16: Scalability comparison between direct temporal-history prompting and GraphRAG retrieval as the accumulated scene history (temporal graphs) increases.*

The scalability advantage of this memory formulation is illustrated in Figure 16, which compares direct temporal-history prompting with GraphRAG retrieval as accumulated scene history increases. In the direct-history approach, prompt length and prediction latency grow with operational duration until the Qwen2.5-3B-Instruct context window of 131,072 tokens is reached at approximately 43 history entries, after which additional history cannot be explicitly incorporated without truncation. In contrast, GraphRAG retrieves only semantically relevant historical graph summaries, keeping prompt size bounded at approximately 15k tokens and prediction latency near 10 s despite continued growth in stored history. This bounded inference behaviour avoids context-window saturation and supports scalable longitudinal reasoning for continuous underground operation.

### 3.2.5 Failure Case Analysis

A failure case analysis was conducted to understand the conditions under which the reasoning layers failed, rather than reporting only aggregate hazard coverage. The observed failures were associated with three distinct sources: structured-output instability, reasoning limitations in the LLM reasoning layer, and reasoning limitations in the GraphRAG-based longitudinal layer, as summarised in Table 6.

*Table 6: Summary of failure cases in the reasoning layers*

| Layer | Number of cases | Main affected hazard type | Likely cause |
|---|---|---|---|
| Reasoning pipeline | 3 | Structured-output failure | Structured outputs did not fully conform to the required JSON schema and could not be reliably parsed by the downstream program |
| LLM reasoning layer | 1 | unmodelled structure near operational path | Anomaly interpreted as a low-priority scene irregularity |
| Longitudinal reasoning layer | 1 | Progressive structural deterioration | Gradual deformation trends were insufficiently preserved in memory summaries for reliable escalation reasoning |
| Longitudinal reasoning layer | 1 | Structural anomaly recurring at the same spatial location | Persistent anomaly recurrence was interpreted as stable background variation rather than progressive hazard escalation |

| Longitudinal reasoning layer | 1 | Equipment behaviour deviation | Irregular motion patterns were interpreted as operational variation rather than abnormal behaviour |
|---|---|---|---|
| Longitudinal reasoning layer | 1 | Recurring near-miss interaction | Repeated sub-threshold interactions were underweighted in cumulative risk reasoning |

The failure cases reveal that the remaining errors stem from identifiable limits in the reasoning pipeline rather than from random model behaviour. One class of failures was associated with structured-output robustness rather than hazard interpretation itself. In three cases, the model produced responses that could not be reliably parsed by the downstream pipeline, even after two retry attempts; these responses were then treated as output failures. The single standalone LLM reasoning miss suggests that weak anomaly evidence, without supporting threshold violations, can be underweighted when assigning hazard significance based solely on the current temporal graph. In contrast, the unresolved GraphRAG cases were concentrated in hazards characterised by gradual, low-gradient changes, whose safety signals accumulate over extended histories. These cases highlight a limitation in the current memory representation. Although graph summaries preserve relational and categorical context, they may not retain enough quantitative detail to distinguish slow escalation from normal variation.

Taken together, the failure analysis points to two priorities for future improvement. First, stronger structured-output enforcement is needed to improve downstream reliability. Second, the memory representation should be extended to retain compact quantitative trend descriptors, such as deformation progression, repeated near-miss frequency, and other cumulative temporal indicators, alongside the existing qualitative graph summaries. Overall, the remaining errors appear to arise from identifiable limits in representation, retrieval, and output robustness, rather than from weaknesses in the broader reasoning framework.

### 3.3 End-to-End Pipeline Inference Performance

To assess real-time feasibility, end-to-end latency was measured across all pipeline stages on a workstation equipped with an NVIDIA GeForce RTX 4080 (16 GB VRAM). The results are summarised in Table 7.

*Table 7: End-to-end latency breakdown of the proposed multi-layer hazard reasoning pipeline*

| Pipeline Level | Stage | Latency (ms) |
|---|---|---|
| Frame-level pipeline (input interval ~250 ms, 4 Hz) | Voxelisation and sparse tensor construction | 8.3 |
| | Sparse 3D perception (forward pass) | 32.9 |
| | Uncertainty map computation and anomaly clustering | 5.8 |
| | Scene graph construction with node attributes | 24.2 |
| | Temporal graph update and kinematic extraction | 14.1 |
| | Deterministic rule evaluation | < 1.0 |
| Total frame-level latency | | **86.3** |
| Temporal graph level (LLM reasoning) Runs every 10 seconds | LLM first-token latency (Qwen2.5-3B-Instruct) | 123.5 |
| | LLM full generation (contextual reasoning) | 6601.8 |
| Total LLM assessment latency | | **6725.3** |
| GraphRAG level Triggered based | Graph summary embedding (Qwen3-Embedding-0.6B) | 85.9 |
| | Qdrant vector search and metadata filtering | 2.3 |
| | Candidate reranking (Qwen3-Reranker-0.6B) | 22 |
| | GraphRAG-augmented LLM reasoning | 9602.1 |
| Total GraphRAG additional latency | | **9712.3** |

The monitoring devices capture point clouds at 10 Hz. Following SLAM reconstruction and input validation, the processed data enters the decision-making pipeline at 4 Hz, corresponding to one frame every 250 ms. The frame-level pipeline completes processing in 86.3 ms per frame (approximately 12 FPS), which is well within the available 250 ms interval. This confirms that real-time operation is maintained with sufficient computational headroom. Deterministic rule evaluation introduces negligible latency, ensuring that safety-critical alerts are generated within the same processing window.

The LLM reasoning module operates on a temporal graph constructed over a 10-second observation window; its ~6.7 s latency remains within the update interval and therefore does not create a computational bottleneck. In practice, each inference completes before the next temporal reasoning cycle is triggered, allowing the module to operate continuously without delaying frame-level perception or deterministic safety alerts. GraphRAG is invoked only when additional historical context is required, and its latency affects only the corresponding temporal reasoning update, rather than the continuous frame-level hazard-monitoring pipeline.

## 4. Discussion

The results show that the proposed framework improves underground monitoring at both the perception and reasoning levels. At the perception stage, the sparse Minkowski UNet with contrastive pretraining achieves 0.86 mIoU and 92.7% overall accuracy, outperforming both the widely used PointNet++ baseline and the more recent Point Transformer V3 model, while maintaining 30.2 FPS with only 94 MB of GPU memory. Unlike approaches optimised for segmentation accuracy alone, the architecture was designed to balance accuracy, computational efficiency, and practical deployability under the constraints of underground edge computing. The sparse voxel formulation avoids unnecessary computation in empty tunnel space, preserving memory and bandwidth for higher-level reasoning layers, which is as important as raw segmentation performance for a system intended for continuous operation.

These results should be understood in the context of the dataset design. The combination of real roadway scans, scan-and-merge object placement, and a high-fidelity Gazebo simulator was not merely a data-augmentation choice but a necessary response to the limited availability of annotated underground longwall data. The scan-and-merge approach enables realistic object placement and preserves authentic sensor characteristics from limited field access, while the simulator contributes temporally rich dynamic scenarios that cannot be safely collected in live production panels. Likewise, the benefit of contrastive pretraining should be understood as a data-efficiency mechanism rather than purely a performance gain. The sim-to-real experiment makes this concrete: training on simulated data alone yields approximately 0.21 mIoU on real scans, whereas incorporating just 16 annotated real underground point clouds within the two-stage training strategy raises performance to 0.79 mIoU. Together, these design choices provide a replicable pathway for building practical 3D perception systems in environments where data scarcity is a structural constraint rather than a temporary limitation.

The entropy-based anomaly features embedded in the scene graph serve a role beyond uncertainty quantification. By propagating per-voxel predictive entropy into the graph representation, the framework converts uncertain or unfamiliar regions into structured evidence that can be interpreted by the reasoning layer. This is particularly important in underground mining, where hazardous conditions cannot be exhaustively defined in advance, and structural degradation may emerge without clear categorical labels. The roof bolt identification in the real underground roadway scenes illustrates this capability. Roof bolts were not included as a labelled class, yet their geometric irregularity produced spatially clustered high-entropy anomaly nodes along the roof surface. Qwen2.5-3B-Instruct interpreted the spatial distribution of these clusters as roof support infrastructure and flagged a localised region of reduced bolt density for further inspection as a possible structural support deficiency. This potential issue was not part of any predefined scenario and was inferred entirely from the anomaly pattern represented in the scene graph. This shows how uncertain sensory observations can be transformed into explicit and traceable knowledge elements, allowing the framework to interpret previously unmodelled conditions and incorporate them into subsequent safety reasoning.

The layered reasoning architecture reflects a deliberate design response to the heterogeneous nature of underground hazards. The deterministic rule engine correctly identified all 65 within-scope scenarios with zero false positives, confirming that explicit safety constraints such as proximity violations, TTC breaches, and visibility degradation can be enforced reliably at the frame level. This result also has practical implications for safety assurance. Because the deterministic layer produces explicit threshold-defined outputs, its decisions can be inspected, documented, and aligned with site-specific safety rules more readily than probabilistic model outputs. This makes the rule-based layer suitable as the authoritative safety gate within the proposed architecture, while also preserving a clear audit trail through graph-level safety flags.

However, many safety-critical hazards are not initially expressed as direct threshold violations. They may emerge from spatial context, interaction patterns, abnormal behaviour, repeated exposure, or anomaly evidence that requires higher-level interpretation. The LLM reasoning layer addresses this gap by resolving 22 of the 25 contextual scenarios identifiable from the temporal graph alone, increasing overall hazard coverage from 57% to 76%. GraphRAG-based longitudinal reasoning further strengthens this by resolving 20 of the remaining 25 history-dependent hazards, increasing overall coverage to 93%. The largest gains occur in recurring interactions and progressive hazard patterns that appear ambiguous within a short observation window but become identifiable when interpreted against relevant historical graph states. Importantly, this improvement is achieved through selective retrieval rather than full-history prompting, keeping prompt size bounded and reducing context dilution over long operational sequences.

These results also clarify both the capability and the boundary of the proposed reasoning approach. While most hazards are effectively captured through contextual and longitudinal reasoning, those driven by gradual quantitative changes remain more challenging. Graph-based summaries preserve relational context efficiently, but are less sensitive to incremental deformation, repeated sub-threshold interactions, or minor motion irregularities that accumulate over time. This suggests that extending the memory representation to retain compact temporal trend descriptors alongside qualitative graph summaries would improve sensitivity to low-gradient hazard evolution.

At a system level, this distinction also defines how language-model-based reasoning should be positioned in safety-critical underground monitoring. The deterministic rule layer is best suited for enforcing explicit safety constraints, since its outputs are threshold-defined, auditable, and directly traceable to predefined safety conditions. In contrast, the LLM and GraphRAG layers provide contextual and longitudinal interpretation rather than certified control authority. Their outputs are evidence-referenced through graph nodes, edges, and retrieved memories, which supports post-hoc review and operator understanding; however, their probabilistic and context-dependent nature makes them less suitable as independent safety-critical actuators. The appropriate deployment role of these layers is therefore decision support: they extend operator awareness by identifying complex relational and temporal hazards, while deterministic rules remain responsible for immediate safety enforcement. Future integration with mine safety management systems should make this hierarchy explicit by separating certified, rule-based alerts from advisory contextual reasoning within the system architecture and safety case documentation.

Beyond technical performance, it is important to examine the practical barriers that may govern real-world adoption of the proposed framework. First, any sensing or computing hardware used in explosion-risk underground atmospheres must satisfy flameproof or intrinsically safe certification requirements. This requirement is addressed through the broader monitoring platform on which the proposed framework is built: a compact, thermally stable, and continuously operating LiDAR–camera sensing system designed for on-device processing and reliable underground data communication. The platform has been tested and certified in accordance with IEC 60079-0:2017 and IEC 60079-1:2014, supporting safe deployment in methane-hazardous underground coal mine environments [12, 57]. Second, operator trust remains essential: mine personnel must understand why an alert is generated before relying on it in safety-critical contexts. The graph-serialised and evidence-referenced output format used in this work is intended to support such trust, but structured human-factors evaluation with mine operators and safety personnel remains necessary. Third, the GraphRAG memory archive stores structured summaries of operational context, including personnel locations, equipment states, interaction patterns, and inferred risk conditions. To support data control, the framework is designed to run within the mine's local computing environment, using on-device LLM inference and locally hosted graph-memory storage rather than external cloud services. This keeps graph memories, prompts, and model outputs within the mine's own infrastructure while still requiring appropriate governance, access controls, retention policies, and audit procedures prior to operational deployment. These considerations highlight the practical issues that need to be addressed when translating the framework from controlled research setting to operational mine-site use.

Overall, the proposed framework should be understood as an architectural pathway for moving underground mine monitoring from 3D perception to safety reasoning, rather than as a fixed system configuration. The study shows that colourised 3D sensing, data-efficient semantic perception, uncertainty-aware anomaly detection, graph-based representation, deterministic rules, contextual LLM reasoning, and GraphRAG-based longitudinal memory can operate together within a single continuous pipeline. The specific language, embedding, and retrieval models used here represent one practical implementation under current on-device hardware constraints, and these components can be replaced as compact AI models improve without changing the core architecture. At the same time, broader validation in natural underground operation remains an important next step, as real deployments may involve sensor drift, intermittent communication, unexpected object configurations, changing dust and lighting conditions, and slowly accumulating perception errors over long durations. In this sense, the main contribution is a modular and extensible methodology that links real-time underground sensing with structured reasoning and memory, providing a foundation for future intelligent mine-monitoring decision-support systems that can evolve with advances in locally deployed AI and operational practice.

## 5.

## 6. Conclusion

This paper presented a unified perception-to-reasoning framework for autonomous safety monitoring in underground mining, combining 3D perception and graph-based reasoning within a single operational pipeline. The results show that this approach extends monitoring beyond conventional detection-based systems toward more context-aware safety interpretation. The key findings of the study are as follows:

1. The study developed and evaluated a complete monitoring framework that connects colourised 3D sensing, semantic perception, graph-based representation, rule-based checks, on-device LLM reasoning, and historical retrieval within one continuous pipeline. This demonstrates that sensing, perception, and reasoning can be integrated without treating them as separate monitoring stages.

2. The study established a practical data strategy for restricted underground environments by combining real roadway scans, controlled object placement, and simulated longwall scenes. This provides a realistic pathway for developing and testing 3D perception systems when repeated access to active production panels is not feasible.

3. The data-efficient 3D perception approach improved segmentation performance under limited labelled data. Contrastive self-supervised pretraining followed by supervised fine-tuning achieved 0.86 mIoU and 92.7% overall accuracy while operating at 30.2 FPS with a 94 MB memory footprint.

4. The uncertainty-aware anomaly detection method helped identify hazards beyond predefined semantic classes. By using per-voxel predictive entropy, the system located uncertain regions and generated anomaly proposals, with qualitative results showing that these cues can support interpretation of previously unlabelled structural conditions in real underground scenes.

5. The scene and temporal graph representation provided a compact bridge between perception outputs and safety reasoning. Detected objects and anomaly proposals were represented as graph nodes and linked across time to capture distance, proximity, uncertainty, motion, and evolving interactions.

6. The hybrid reasoning layer improved both current and historical hazard interpretation. Rule-based checks detected explicit safety violations, on-device LLM reasoning increased hazard coverage from 57% to 76%, and GraphRAG-based retrieval further increased overall coverage to approximately 93% across 115 evaluated scenarios.

Overall, the study demonstrates that structured 3D perception combined with graph-based reasoning and on-device language models provides a viable foundation for continuous, intelligent, and context-aware underground safety monitoring, with the potential to meaningfully reduce operator burden and make critical hazards less likely to go unnoticed in active underground mining environments.

## CRediT authorship contribution statement

**Pasindu Ranasinghe:** Writing – original draft, Data curation, Investigation, Conceptualisation, Methodology, Software, Formal analysis, Visualisation, Validation. **Simit Raval:** Writing – review and editing, Conceptualisation, Project administration, Funding acquisition, Resources, Supervision. **Dibyayan Patra:** Writing – review and editing, Conceptualisation, Formal analysis, Validation. **Bikram Banerjee:** Writing – review and editing, Validation, Supervision. **Ismet Canbulat:** Writing – review and editing, Validation, Supervision.

## Funding

This project is supported by the Key Science and Technology Innovation Project of China Coal Technology & Engineering Group Corp under Grant No. 2022-3-KJHZ005, and was conducted in collaboration with Beijing Tianma Intelligent Control Technology Co., Ltd.

Declaration of competing interest

The authors declare that they have no known competing financial interests or personal relationships that could have appeared to influence the work reported in this paper.

Data statement

The data supporting this study are not publicly available due to mine operator confidentiality requirements and active data-sharing agreements. The data may be made available by the corresponding author upon reasonable request, subject to the required mine-site approvals.

Acknowledgments

The authors acknowledge the assistance of Kanchana Gamage (Technical Officer) and Mark Whelan (Senior Technical Officer) from the School of Minerals and Energy Resources Engineering, UNSW, for their valuable support during sensor setup, data acquisition, and testing activities associated with this study.

Declaration of generative AI and AI-assisted technologies in the writing process

During the preparation of this work, the authors used OpenAI's ChatGPT to improve the language and readability of the manuscript. After using this tool, the authors reviewed and edited the content as needed and take full responsibility for the content of the published article.